\documentclass[10pt,twocolumn,letterpaper]{article}%
\pdfoutput=1 
\usepackage[table,usenames,dvipsnames]{xcolor}%
\usepackage[utf8]{inputenc}%
\usepackage[T1]{fontenc}%
\usepackage{cvpr}%
\usepackage{times}%
\usepackage{epsfig}%
\usepackage{graphicx}%
\usepackage{amsmath}%
\usepackage{amssymb}%
\usepackage{multirow}%
\usepackage{float}%
\usepackage{tikz}%
\usetikzlibrary{tikzmark}%
\usepackage{pifont}
\usepackage{colortbl}%
\usepackage[normalem]{ulem}%
\usepackage{xspace}%
\usepackage{comment}
\usepackage{anyfontsize}

\newcommand{\pz}{\phantom{0}}%
\newcommand{\Aceil}[1]{\lceil #1 \rceil}%
\newcommand{\Afloor}[1]{\lfloor #1 \rfloor}%

\definecolor{dgreen}{rgb}{0.0,0.6,0.0} 
\definecolor{dred}{rgb}{0.6,0.0,0.0} 
\definecolor{BrickRed}{rgb}{0.72,0.0,0.0}%
\definecolor{alexey}{rgb}{0.7,0,1}%
\definecolor{grey}{rgb}{0.6,0.6,0.6}%

\newcommand{\my}{\textcolor{dgreen}{\ding{51}}}%
\newcommand{\mn}{\textcolor{dred}{\ding{55}}}%

\newcommand{\sinc}[1]{\textcolor{blue}{#1}}	 
	 
\newcommand{\kitc}[1]{\textcolor{dgreen}{#1}}	 

\newcommand{\chairsSD}{\mbox{ChairsSDHom}\xspace}	 
\newcommand{\chairs}{\mbox{Chairs}\xspace}	 
\newcommand{\things}{\mbox{Things3D}\xspace}	 
\newcommand{\FN}[1]{\mbox{FlowNet2-#1}\xspace}	 
\newcommand{\Sshort}{S_\mathit{short}}
\newcommand{\Slong}{S_\mathit{long}}
\newcommand{\Sfine}{S_\mathit{fine}}
\newcommand{\neta}{Net1\xspace}
\newcommand{\netb}{Net2\xspace}

\def\cvprPaperID{900} 

\cvprfinalcopy 

\begin{document}

\title{FlowNet 2.0: Evolution of Optical Flow Estimation with Deep Networks}

\newcommand{\authorfont}[1]{#1}

\author{
\authorfont{Eddy Ilg},\,\, 
\authorfont{Nikolaus Mayer},\,\,
\authorfont{Tonmoy Saikia},\,\,
\authorfont{Margret Keuper},\,\, 
\authorfont{Alexey Dosovitskiy},\,\, 
\authorfont{Thomas Brox} \vspace*{1mm}
\\
\authorfont{University of Freiburg, Germany}\\
{\tt\small \{ilg,mayern,saikiat,keuper,dosovits,brox\}@cs.uni-freiburg.de}
}

\maketitle

\begin{abstract}
The FlowNet demonstrated that optical flow estimation can be cast as a learning problem. However, the state of the art with regard to the quality of the flow has still been defined by traditional methods. Particularly on small displacements and real-world data, FlowNet cannot compete with variational methods. In this paper, we advance the concept of end-to-end learning of optical flow and make it work really well.  
The large improvements in quality and speed are caused by three major contributions: first, we focus on the training data and show that the schedule of presenting data during training is very important. Second, we develop a stacked architecture that includes warping of the second image with intermediate optical flow.
Third, we elaborate on small displacements by introducing a sub-network specializing on small motions.
FlowNet~2.0 is only marginally slower than the original FlowNet but decreases the estimation error by more than 50\%. 
It performs on par with state-of-the-art methods, while running at interactive frame rates. 
Moreover, we present faster variants that allow optical flow computation at up to 140fps with accuracy matching the original FlowNet. 
\end{abstract}

\section{Introduction}
\label{sec:introduction}

The FlowNet by Dosovitskiy \etal~\cite{DFIB15} represented a paradigm shift in optical flow estimation. The idea of using a simple convolutional CNN architecture to directly learn the concept of optical flow from data was completely disjoint from all the established approaches. However, first implementations of new ideas often have a hard time competing with highly fine-tuned existing methods, and FlowNet was no exception to this rule. It is the successive consolidation that resolves the negative effects and helps us appreciate the benefits of new ways of thinking.

At the same time, it resolves problems with small displacements and noisy artifacts in estimated flow fields.
This leads to a dramatic performance improvement on real-world applications such as action recognition and motion segmentation, bringing FlowNet 2.0 to the state-of-the-art level.

\begin{figure}
  \begin{center}
    \includegraphics[width=\linewidth]{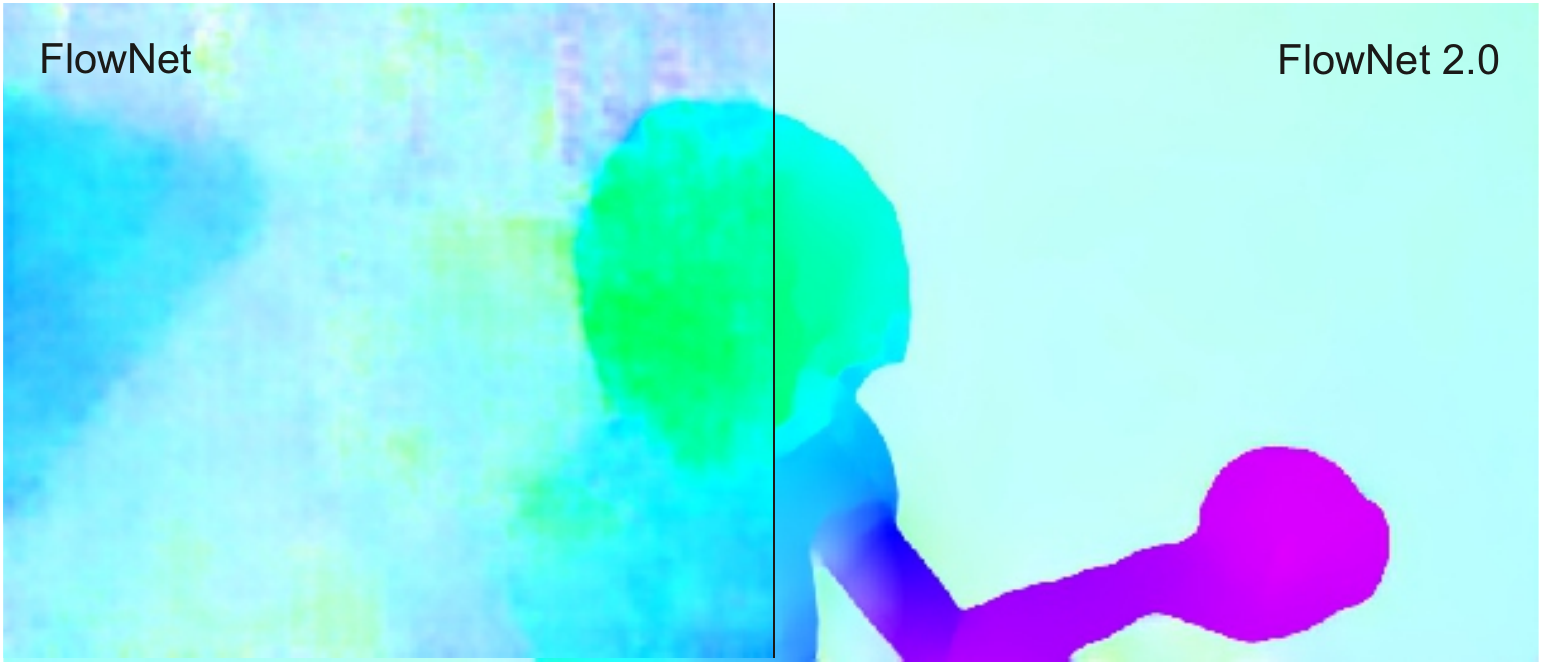}
  \end{center}
  \caption{We present an extension of FlowNet. FlowNet~2.0 yields smooth flow fields, preserves fine motion details and runs at $8$ to $140$fps. The accuracy on this example is four times higher than with the original FlowNet.} 
\label{fig:teaser}
\end{figure}

\begin{figure*}[t]
  \begin{center}
      \includegraphics[width=\linewidth]{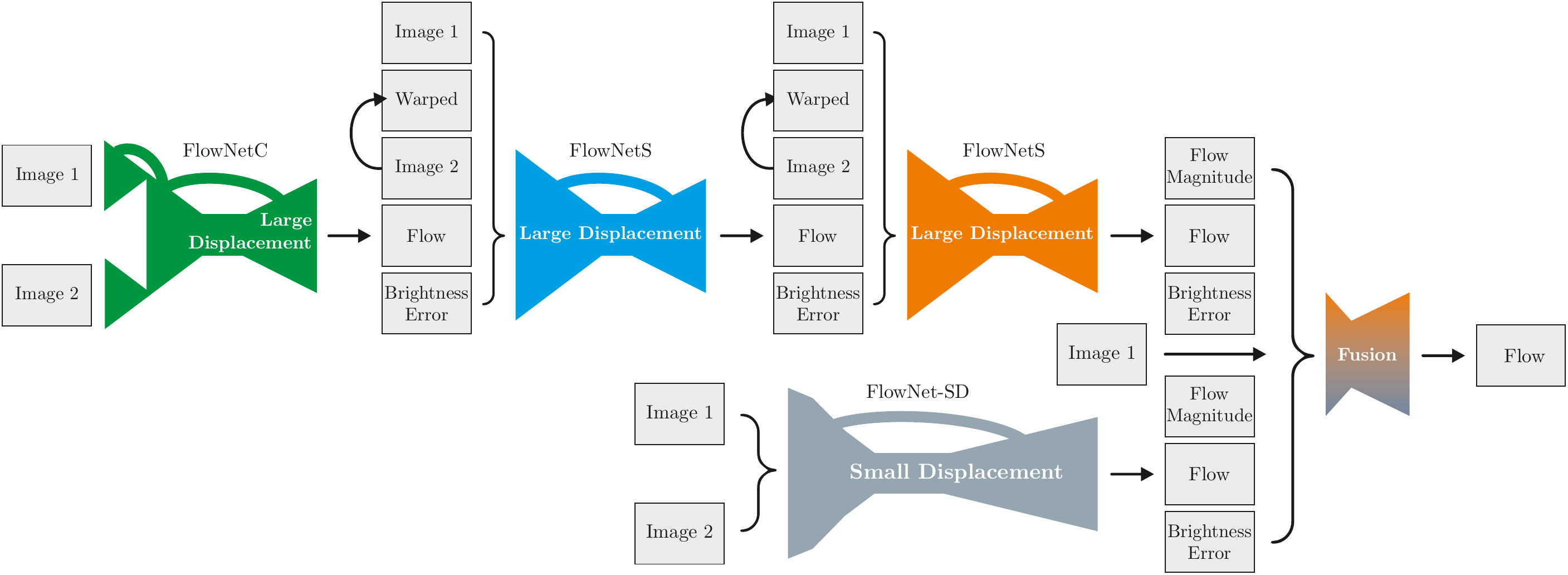}
  \end{center}
  \caption{Schematic view of complete architecture: To compute large displacement optical flow we combine multiple FlowNets. Braces indicate concatenation of inputs. \textit{Brightness Error} is the difference between the first image and the second image warped with the previously estimated flow. 
  To optimally deal with small displacements, we introduce smaller strides in the beginning and convolutions between upconvolutions into the FlowNetS architecture. Finally we apply a small fusion network to provide the final estimate. 
  }
  \label{fig:schematic}
\end{figure*}%

The way towards FlowNet~2.0 is via several evolutionary, but decisive modifications that are not trivially connected to the observed problems. First, we evaluate the influence of dataset schedules. Interestingly, the more sophisticated training data provided by Mayer \etal~\cite{MIFDB16} leads to inferior results if used in isolation. However, a learning schedule consisting of multiple datasets improves results significantly. In this scope, we also found that the FlowNet version with an explicit correlation layer outperforms the version without such layer. This is in contrast to the results reported in Dosovitskiy \etal~\cite{DFIB15}. 

As a second contribution, we introduce a warping operation and show how stacking multiple networks using this operation can significantly improve the results. 
By varying the depth of the stack and the size of individual components we obtain
many network variants with different size and runtime. This allows us to control the trade-off between accuracy and computational resources. We provide networks for the spectrum between $8$fps and $140$fps. 

Finally, we focus on small, subpixel motion and real-world data. To this end, we created a special training dataset and a specialized network.
We show that the architecture trained with this dataset performs well on small motions typical for real-world videos. 
To reach optimal performance on arbitrary displacements, we add a network that learns to fuse the former stacked network with the small displacement network in an optimal manner.   

The final network outperforms the previous FlowNet by a large margin and performs on par with state-of-the-art methods on the Sintel and KITTI benchmarks. 
It can estimate small and large displacements with very high level of detail while providing interactive frame rates. 

%

\section{Related Work} 
End-to-end optical flow estimation with convolutional networks was proposed by Dosovitskiy \etal in~\cite{DFIB15}. Their model, dubbed FlowNet, takes a pair of images as input and outputs the flow field. 
Following FlowNet, several papers have studied optical flow estimation with CNNs: featuring a 3D convolutional network~\cite{Tran2016voxel}, an unsupervised learning objective~\cite{Ahmadi2016unsupflow,Yu2016backtobasics}, carefully designed rotationally invariant architectures~\cite{Teney2016cnnflow}, or a pyramidal approach based on the coarse-to-fine idea of variational methods~\cite{SpyNet}. None of these methods significantly outperforms the original FlowNet.

An alternative approach to learning-based optical flow estimation is to use CNNs for matching image patches.
Thewlis \etal~\cite{Thewlis2016trainabledm} formulate Deep Matching~\cite{deepflow} as a convolutional network and optimize it end-to-end. Gadot~\& Wolf \cite{Gadot2016patchbatch} and Bailer \etal~\cite{Bailer2016match} learn image patch descriptors using Siamese network architectures.
These methods can reach good accuracy, but require exhaustive matching of patches. 
Thus, they are restrictively slow for most practical applications.
Moreover, patch based approaches lack the possibility to use the larger context of the whole image
because they operate on small image patches.

Convolutional networks trained for per-pixel prediction tasks often produce noisy or blurry results. 
As a remedy, out-of-the-box optimization can be applied to the network predictions as a postprocessing operation, for example, optical flow estimates can be refined with a variational approach~\cite{DFIB15}.
In some cases, this refinement can be approximated by neural networks: Chen~\& Pock ~\cite{Chen2016tnrd} formulate reaction diffusion model as a CNN and apply it to image denoising, deblocking and superresolution.
Recently, it has been shown that similar refinement can be obtained by stacking several convolutional networks on top of each other.
This led to improved results in human pose estimation~\cite{NYD16,Carreira2016pose} and semantic instance segmentation~\cite{Romera2016instance}.
In this paper we adapt the idea of stacking multiple networks to optical flow estimation. 

Our network architecture includes warping layers that compensate for some already estimated preliminary motion in the second image. The concept of image warping is common to all contemporary variational optical flow methods and goes back to the work of Lucas~\& Kanade~\cite{LucasKanade}. In Brox \etal~\cite{Brox2004flow} it was shown to correspond to a numerical fixed point iteration scheme coupled with a continuation method. 

The strategy of training machine learning models on a series of gradually increasing tasks is known as curriculum learning~\cite{Bengio2009curriculum}. The idea dates back at least to Elman~\cite{Elman93startingsmall}, who showed that both the evolution of tasks and the network architectures can be beneficial in the language processing scenario. In this paper we revisit this idea in the context of computer vision and show how it can lead to dramatic performance improvement on a complex real-world task of optical flow estimation.


\section{Dataset Schedules \label{sec:dataset-schedules}}

High quality training data is crucial for the success of supervised training.
We investigated the differences in the quality of the estimated optical flow depending on the presented training data. Interestingly, it turned out that not only the kind of data is important but also the order in which it is presented during training. 

The original FlowNets~\cite{DFIB15} were trained on the FlyingChairs dataset (we will call it \chairs). This rather simplistic dataset contains about $22$k image pairs of chairs superimposed on random background images from Flickr. Random affine transformations are applied to chairs and background to obtain the second image and ground truth flow fields. The dataset contains only planar motions. 

The FlyingThings3D (\things) dataset proposed by Mayer \etal~\cite{MIFDB16} can be seen as a three-dimensional version of the FlyingChairs.
The dataset consists of $22$k renderings of random scenes showing 3D models from the ShapeNet dataset~\cite{shapenet} moving in front of static 3D backgrounds. In contrast to \chairs, the images show true 3D motion and lighting effects and there is more variety among the object models. 


We tested the two network architectures introduced by Dosovitskiy \etal~\cite{DFIB15}: FlowNetS, which is a straightforward encoder-decoder architecture, and FlowNetC, which includes explicit correlation of feature maps.
We trained FlowNetS and FlowNetC on \chairs and \things and an equal mixture of samples from both datasets using the different learning rate schedules shown in Figure~\ref{fig:lr-schedules}. The basic schedule $\Sshort$ ($600$k iterations) corresponds to Dosovitskiy \etal~\cite{DFIB15} except some minor changes\footnote{(1) We do not start with a learning rate of $1e-6$ and increase it first, but we start with $1e-4$ immediately. (2) We fix the learning rate for $300$k iterations and then divide it by $2$ every $100$k iterations.}. 
Apart from this basic schedule $\Sshort$, we investigated a longer schedule $\Slong$ with $1.2$M iterations, and a schedule for fine-tuning $\Sfine$ with smaller learning rates.
Results of networks trained on \chairs and \things with the different schedules are given in Table~\ref{tab:schedules-and-datasets}. The results lead to the following observations: 

\begin{figure}
  \begin{center}
      \includegraphics[width=\linewidth]{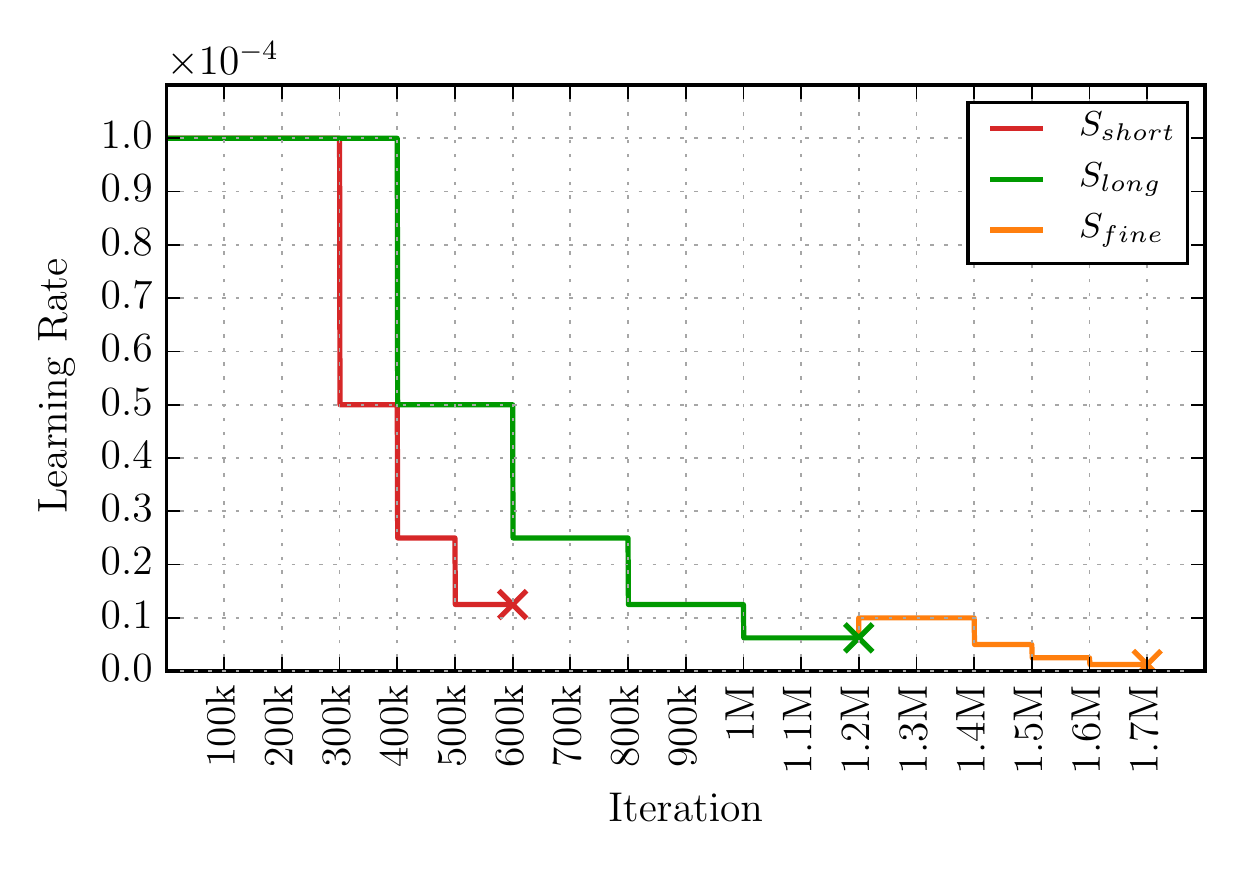}
  \end{center}
  \caption{Learning rate schedules: $\Sshort$ is similar to the schedule in Dosovitskiy \etal~\cite{DFIB15}. We investigated another longer version $\Slong$ and a fine-tuning schedule $\Sfine$.
  }
  \label{fig:lr-schedules}
\end{figure}%

\newcommand{\chairsToThings}{\mbox{\chairs$\!\to$\things}\xspace}%

\begin{table}[t]
  \begin{center}
  {\small
  \setlength{\tabcolsep}{0.1cm}
  \begin{tabular}{|c|c||c||cc|} 
  \hline
  Architecture                       & Datasets               & $\Sshort$ & $\Slong$ & $\Sfine$ \\
  
  \hline
  \hline 
  
  \multirow{5}{*}{FlowNetS}   & \chairs                & $4.45$      & -          &  -         \\
                              & \chairs                 & -           & $4.24$     &  $4.21$    \\
                              & \things                & -           & $5.07$     &  $4.50$    \\
                              & mixed                  & -           & $4.52$     &  $4.10$    \\
                              & \chairsToThings    & -           & $4.24$     &  $\mathbf{3.79}$    \\
                              
  \hline \hline
  
  \multirow{2}{*}{FlowNetC}   & \chairs                & $3.77$      & -          &  -         \\
                              & \chairsToThings    & -           & $3.58$     & \cellcolor{gray!15}$\mathbf{3.04}$    \\   
  \hline
  \end{tabular}
  }
  \end{center}
  \caption{
  Results of training FlowNets with different schedules on different datasets (one network per row). Numbers indicate endpoint errors on Sintel train \textit{clean}. \emph{mixed} denotes an equal mixture of \chairs and \things. Training on \chairs first and fine-tuning on \things yields the best results (the same holds when testing on the KITTI dataset; see supplemental material). FlowNetC performs better than FlowNetS. 
  \label{tab:schedules-and-datasets}
  }
\end{table}

\begin{table*}[t]
  \begin{center}
  \begin{tabular}{|l||c|c|c|c|c|c||c|c|}
  \hline
  Stack                     & \multicolumn{2}{c|}{Training}  & Warping  & Warping  & \multicolumn{2}{c||}{Loss after}  & EPE on \chairs & EPE on Sintel  \\
  architecture              & \multicolumn{2}{c|}{enabled}  & included & gradient & \multicolumn{2}{c||}{}     & test      & train \textit{clean}    \\
  \cline{2-3}
  \cline{6-7}
                            & \neta     & \netb             &          & enabled  & \neta     & \netb                  &           &     \\
                            
  \hline \hline 
  
  \neta                     & \my       & --                & --       & --       & \my       & --                      & $3.01$   & $3.79$ \\
  \neta $+$ \netb           & \mn       & \my               & \mn      & --       & --        & \my                     & $2.60$   & $4.29$ \\
  \neta $+$ \netb           & \my       & \my               & \mn      & --       & \mn       & \my                     & $2.55$   & $4.29$ \\
  \neta $+$ \netb           & \my       & \my               & \mn      & --       & \my       & \my                     & $2.38$   & $3.94$ \\
  \rowcolor{gray!15}
  \neta $+$ W $+$ \netb     & \mn       & \my               & \my      & --       & --        & \my                     & $1.94$   & $\textbf{2.93}$ \\
  \neta $+$ W $+$ \netb     & \my       & \my               & \my      & \my      & \mn       & \my                     & $1.96$   & $3.49$ \\
  \neta $+$ W $+$ \netb     & \my       & \my               & \my      & \my      & \my       & \my                     & $\textbf{1.78}$ & $3.33$ \\
  \hline
  \end{tabular} 
  \end{center}
  \caption{
  Evaluation of options when stacking two FlowNetS networks (\neta and \netb). \neta was trained with the \chairsToThings schedule from Section~\ref{sec:dataset-schedules}. \netb is initialized randomly and subsequently, \neta and \netb together, or only \netb is trained on \chairs with $\Slong$; see text for details. When training without warping, the stacked network overfits to the \chairs dataset. The best results on Sintel are obtained when fixing \neta and training \netb with warping.  
  \label{tab:two_stack_results}
  } 
\end{table*}

\textbf{The order of presenting training data with different properties matters.} Although \things is more realistic, training on \things alone leads to worse results than training on \chairs. The best results are consistently achieved when first training on \chairs and only then fine-tuning on \things. This schedule also outperforms training on a mixture of \chairs and \things. 
We conjecture that the simpler \chairs dataset helps the network learn the general concept of color matching without developing possibly confusing priors for 3D motion and realistic lighting too early. 
The result indicates the importance of training data schedules for avoiding shortcuts when learning generic concepts with deep networks. 

\textbf{FlowNetC outperforms FlowNetS.} The result we got with FlowNetS and $\Sshort$ corresponds to the one reported in Dosovitskiy \etal~\cite{DFIB15}. However, we obtained much better results on FlowNetC. We conclude that Dosovitskiy \etal~\cite{DFIB15} did not train FlowNetS and FlowNetC under the exact same conditions. When done so, the FlowNetC architecture compares favorably to the FlowNetS architecture. 

\textbf{Improved results. }Just by modifying datasets and training schedules, we improved the FlowNetS result reported by Dosovitskiy \etal~\cite{DFIB15} by $\sim 25\%$ and the FlowNetC result by $\sim 30\%$. 

In this section, we did not yet use specialized training sets for specialized scenarios. The trained network is rather supposed to be generic and to work well in various scenarios. An additional optional component in dataset schedules is fine-tuning of a generic network to a specific scenario, such as the driving scenario, which we show in Section~\ref{sec:experiments}.

\section{Stacking Networks} 
\label{sec:network-schedules}

\subsection{Stacking Two Networks for Flow Refinement\label{sec:two_stack}}

All state-of-the-art optical flow approaches rely on iterative methods \cite{ldof,deepflow,epicflow,flowfields}. Can deep networks also benefit from iterative refinement? To answer this, we experiment with stacking multiple FlowNetS and FlowNetC architectures. 

The first network in the stack always gets the images $I_1$ and $I_2$ as input. Subsequent networks get $I_1$, $I_2$, and the previous flow estimate $w_i=(u_i,v_i)^\top$, where $i$ denotes the index of the network in the stack. 

To make assessment of the previous error and computing an incremental update easier for the network, we also
optionally warp the second image $I_2(x,y)$ via the flow $w_i$ and bilinear interpolation to $\tilde I_{2,i}(x,y)=I_2(x+u_i,y+v_i)$. This way, the next network in the stack can focus on the remaining increment between $I_1$ and $\tilde I_{2,i}$. 
When using warping, we additionally provide $\tilde I_{2,i}$ and the error $e_i=||\tilde I_{2,i}-I_1||$ as input to the next network; see Figure~\ref{fig:schematic}. 
Thanks to bilinear interpolation, the derivatives of the warping operation can be computed (see supplemental material for details).  
This enables training of stacked networks end-to-end. 

Table~\ref{tab:two_stack_results} shows the effect of stacking two networks, the effect of warping, and the effect of end-to-end training. We take the best FlowNetS from Section~\ref{sec:dataset-schedules} and add another FlowNetS on top. The second network is initialized randomly and then the stack is trained on \chairs with the schedule $\Slong$. We experimented with two scenarios: keeping the weights of the first network fixed, or updating them together with the weights of the second network. In the latter case, the weights of the first network are fixed for the first 400k iterations to first provide a good initialization of the second network. 
We report the error on Sintel train \textit{clean} and on the test set of \chairs. Since the \chairs test set is much more similar to the training data than Sintel, comparing results on both datasets allows us to detect tendencies to over-fitting.   

\textbf{We make the following observations:}
(1) Just stacking networks without warping improves results on \chairs but decreases performance on Sintel, i.e. the stacked network is over-fitting. 
(2) With warping included, stacking always improves results. 
(3) Adding an intermediate loss after \neta is advantageous when training the stacked network end-to-end. 
(4) The best results are obtained when keeping the first network fixed and only training the second network after the warping operation. 

Clearly, since the stacked network is twice as big as the single network, over-fitting is an issue. The positive effect of flow refinement after warping can counteract this problem, yet the best of both is obtained when the stacked networks are trained one after the other, since this avoids over-fitting while having the benefit of flow refinement.  


\subsection{Stacking Multiple Diverse Networks}

Rather than stacking identical networks, it is possible to stack networks of different type (FlowNetC and FlowNetS). 
Reducing the size of the individual networks is another valid option. We now investigate different combinations and additionally also vary the network size. 

\begin{figure}
  \begin{center}
      \includegraphics[width=\linewidth]{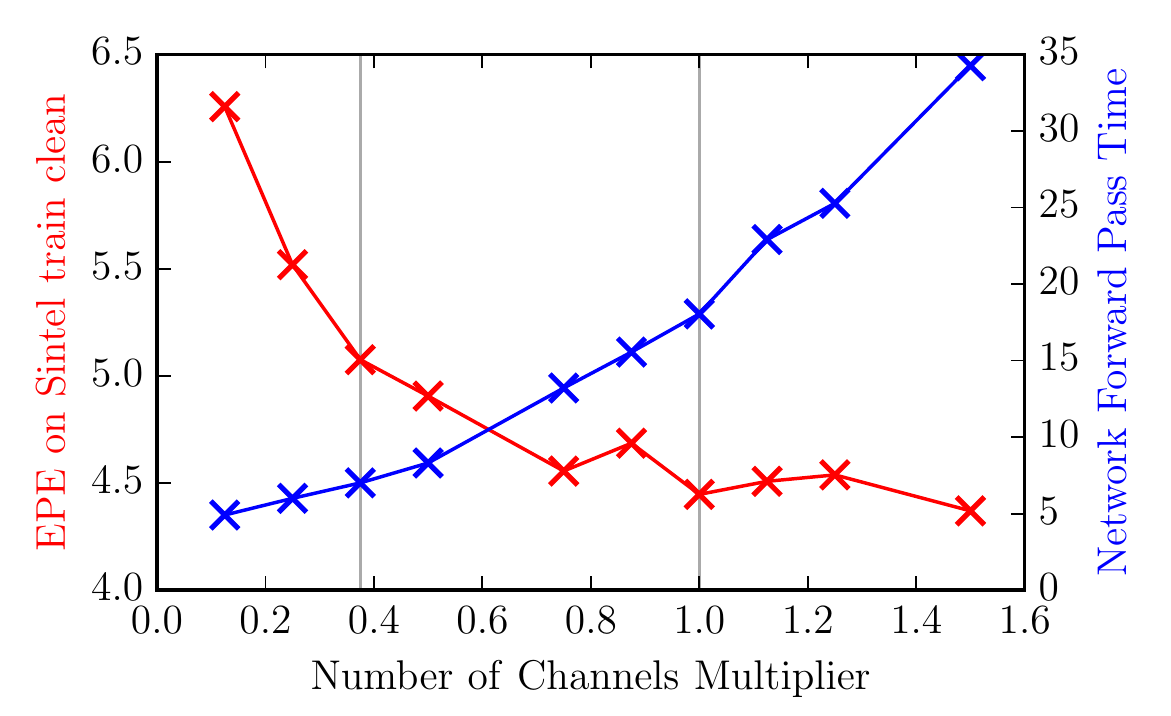}
  \end{center}
  \caption{Accuracy and runtime of FlowNetS depending on the network width. The multiplier 1 corresponds to the width of the original FlowNet architecture. Wider networks do not improve the accuracy. For fast execution times, a factor of $\frac{3}{8}$ is a good choice. Timings are from an Nvidia GTX 1080.
  }
  \label{fig:channel-factors}
\end{figure}

We call the first network the \textit{bootstrap} network as it differs from the second network by its inputs. The second network could however be repeated an arbitray number of times in a recurrent fashion. We conducted this experiment and found that applying a network with the same weights multiple times and also fine-tuning this recurrent part does not improve results (see supplemental material for details). As also done in \cite{NYD16,Chen2016tnrd}, we therefore add networks with different weights to the stack. Compared to identical weights, stacking networks with different weights increases the memory footprint, but does not increase the runtime. In this case the top networks are not constrained to a general improvement of their input, but can perform different tasks at different stages and the stack can be trained in smaller pieces by fixing existing networks and adding new networks one-by-one. We do so by using the \chairsToThings schedule from Section~\ref{sec:dataset-schedules} for every new network and the best configuration with warping from Section~\ref{sec:two_stack}. Furthermore, we experiment with different network sizes and alternatively use FlowNetS or FlowNetC as a bootstrapping network. 
We use FlowNetC only in case of the bootstrap network, as the input to the next network is too diverse to be properly handeled by the Siamese structure of FlowNetC. 
Smaller size versions of the networks were created by taking only a fraction of the number of channels for every layer in the network. Figure \ref{fig:channel-factors} shows the network accuracy and runtime for different network sizes of a single FlowNetS. Factor $\frac38$ yields a good trade-off between speed and accuracy when aiming for faster networks.

\newcommand{\tm}[1]{\textcolor{black}{$#1$ms}}	
\newcommand{\arch}[1]{\textcolor{black}{#1}}	
\newcommand{\nweights}[1]{$#1$}	
\newcommand{\epe}[1]{$#1$}	

\newcommand{\tmlabel}[1]{\textcolor{black}{#1}}	
\newcommand{\archlabel}[1]{\textcolor{black}{#1}}	
\newcommand{\weightslabel}[1]{#1}	
\newcommand{\epelabel}[1]{#1}	

\begin{table}
  \begin{center}
  \resizebox{0.8\linewidth}{!}{%
  \begin{tabular}{|l||c|c|c|c|}%
    \hline
                          & \multicolumn{4}{c|}{Number of Networks} \\
    \cline{2-5}                      
                          & 1  & 2 & 3 & 4 \\
                          
    \hline\hline 
    
    \archlabel{Architecture}     & \arch{s}   & \arch{ss}  & \arch{sss} &  \\ 
    \tmlabel{Runtime}            & $\textbf{7}$\textbf{ms}     & \tm{14}    & \tm{20}    & -- \\ 
    \epelabel{EPE}               & \epe{4.55} & \epe{3.22} & \epe{3.12} &  \\
    \hline
    \archlabel{Architecture}     & \arch{S}   & \arch{SS}  &            &  \\ 
    \tmlabel{Runtime}            & \tm{18}    & \tm{37}    & --         & -- \\ 
    \epelabel{EPE}               & \epe{3.79} & \epe{2.56} &            &  \\
    
    \hline\hline 
    
    \archlabel{Architecture}     & \arch{c}   & \arch{cs}  & \arch{css} & \arch{csss} \\ 
    \tmlabel{Runtime}            & \tm{17}    & \tm{24}    & \tm{31}    & \tm{36} \\ 
    \epelabel{EPE}               & \epe{3.62} & \epe{2.65} & \epe{2.51} & \epe{2.49} \\
    \hline
    \archlabel{Architecture}     & \arch{C}   & \arch{CS}  & \cellcolor{gray!15} \arch{CSS} &  \\ 
    \tmlabel{Runtime}            & \tm{33}    & \tm{51}    & \cellcolor{gray!15} \tm{69}    & -- \\ 
    \epelabel{EPE}               & \epe{3.04} & \epe{2.20} & \cellcolor{gray!15} \epe{\textbf{2.10}} & \\
    \hline
  \end{tabular}
  }
  \end{center}
  \caption{Results on Sintel train \textit{clean} for some variants of stacked FlowNet architectures following the best practices of Section~\ref{sec:dataset-schedules} and Section~\ref{sec:two_stack}. Each new network was first trained on \chairs with $\Slong$ and then on \things with $\Sfine$ (\chairsToThings schedule). Forward pass times are from an Nvidia GTX 1080.  
  \label{tab:flow-net-stacks}
  }
\end{table} 

\noindent
\fbox{\begin{minipage}{0.97\linewidth}
\textbf{Notation:} 
We denote networks trained by the \chairsToThings schedule from Section~\ref{sec:dataset-schedules} starting with \textit{FlowNet2}. Networks in a stack are trained with this schedule one-by-one. For the stack configuration we append upper- or lower-case letters to indicate the original FlowNet or the thin version with $\frac{3}{8}$ of the channels. E.g: \textit{FlowNet2-CSS} stands for a network stack consisting of one FlowNetC and two FlowNetS. \textit{FlowNet2-css} is the same but with fewer channels. 
\end{minipage}}

Table \ref{tab:flow-net-stacks} shows the performance of different network stacks. Most notably, the final FlowNet2-CSS result improves by $\sim 30\%$ over the single network FlowNet2-C from Section \ref{sec:dataset-schedules} and by $\sim 50\%$ over the original FlowNetC~\cite{DFIB15}. Furthermore, two small networks in the beginning always outperform one large network, despite being faster and having fewer weights: FlowNet2-ss ($11$M weights) over FlowNet2-S ($38$M weights), and FlowNet2-cs ($11$M weights) over FlowNet2-C ($38$M weights). Training smaller units step by step proves to be advantageous and enables us to train very deep networks for optical flow. At last, FlowNet2-s provides nearly the same accuracy as the original FlowNet~\cite{DFIB15}, while running at $140$ frames per second. 

\section{Small Displacements\label{sec:small_displacements}} 

\subsection{Datasets}


While the original FlowNet~\cite{DFIB15} performed well on the Sintel benchmark, limitations in real-world applications have become apparent.
In particular, the network cannot reliably estimate small motions (see Figure~\ref{fig:teaser}). This is counter-intuitive, since small motions are easier for traditional methods, and there is no obvious reason why networks should not reach the same performance in this setting. Thus, we examined the training data and compared it to the UCF101 dataset~\cite{ucf101} as one example of real-world data.
While \chairs are similar to Sintel,
UCF101 is fundamentally different (we refer to our supplemental material for the analysis): Sintel
is an action movie and as such contains many fast movements that are difficult for traditional methods, while the displacements we see in the UCF101 dataset are much smaller, mostly smaller than $1$ pixel. Thus, we created a dataset in the visual style of \chairs but with very small displacements and a displacement histogram much more like UCF101. 
We also added cases with a background that is homogeneous or just consists of color gradients. 
We call this dataset \textit{\chairsSD} and will release it upon publication. 

\subsection{Small Displacement Network and Fusion}

We fine-tuned our \FN{CSS} network for smaller displacements by further training the whole network stack on a mixture of \things and \chairsSD and by applying a non-linearity to the error to downweight large displacements\footnote{For details we refer to the supplemental material}. We denote this network by \textit{\mbox{\FN{CSS-ft-sd}}}.
\addtocounter{footnote}{-1}
This increases performance on small displacements and we found that this particular mixture does not sacrifice performance on large displacements. However, in case of subpixel motion, noise still remains a problem and we conjecture that the FlowNet architecture might in general not be perfect for such motion. Therefore, we slightly modified the original FlowNetS architecture and removed the stride $2$ in the first layer. We made the beginning of the network deeper by exchanging the  $7\!\times\!7$ and $5\!\times\!5$ kernels in the beginning with multiple $3\!\times\!3$ kernels\footnotemark. Because noise tends to be a problem with small displacements, we add convolutions between the upconvolutions to obtain smoother estimates as in~\cite{MIFDB16}. We denote the resulting architecture by \textit{\mbox{\FN{SD}}}; see Figure~\ref{fig:schematic}.
\addtocounter{footnote}{-1}

Finally, we created a small network that fuses \FN{CSS-ft-sd} and \FN{SD} (see Figure~\ref{fig:schematic}). The fusion network receives the flows, the flow magnitudes and the errors in brightness after warping as input. It contracts the resolution two times by a factor of $2$ and expands again\footnotemark. Contrary to the original FlowNet architecture it expands to the full resolution. We find that this produces crisp motion boundaries and performs well on small as well as on large displacements. We denote the final network as \textit{FlowNet2}. 

\section{Experiments} 
\label{sec:experiments}

We compare the best variants of our network to state-of-the-art approaches on public bechmarks. In addition, we provide a comparison on application tasks, such as motion segmentation and action recognition. This allows benchmarking the method on real data. 

\subsection{Speed and Performance on Public Benchmarks}

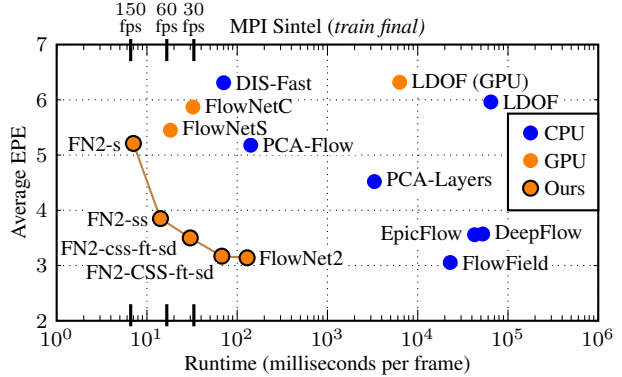
\begin{figure}%
  \begin{center}%
        
    \begin{tikzpicture}[%
      font=\footnotesize,%
      xscale=1.2,yscale=1.75*.42,%
      cpunode/.style={fill=blue,circle,scale=.6},%
      gpunode/.style={fill=orange,circle,scale=.6},%
      ournode/.style={fill=orange,draw,thick,circle,scale=.6},%
      l/.style={fill=white,inner sep=1pt}%
    ]%
      \draw[dotted] (0,0) grid (6,5);%
      \draw[thick] (0,0) -- ++(6,0) -- ++(0,5) -- ++(-6,0) -- cycle;%
      \node at (3,5.3) {MPI Sintel (\textit{train final)}};%
      \node[rotate=90] at (-.4,2.5) {Average EPE};%
      \node at (3,-.8) {Runtime (milliseconds per frame)};%
      \foreach \y in {2,3,...,7} {%
        \node at (0,{\y-2}) [anchor=east] {$\y$};%
        \draw (0,{\y-2}) -- ++(.05,0);%
        \draw (6,{\y-2}) -- ++(-.05,0);%
      }%
      \foreach \x in {0,1,...,6} {%
        \node at (\x,-.3) {$10^{\x}$};%
        \draw (\x,0) -- ++(0,.2);%
        \draw (\x,5) -- ++(0,-.2);%
      }%
      \foreach \logbase in {0,1,...,5} {%
        \foreach \log in {.3,.48,.6,.7,.78,.85,.9,.95} {%
          \draw ({\logbase+\log},0) -- ++(0,.1);%
          \draw ({\logbase+\log},5) -- ++(0,-.1);%
        }%
      }%
      \draw[thick,fill=white] (5,2) rectangle ++(1,1.75);%
      \node[cpunode] at (5.25,3.4) {};%
      \node at (5.3,3.4) [anchor=west] {CPU};%
      \node[gpunode] at (5.25,2.9) {};%
      \node at (5.3,2.9) [anchor=west] {GPU};%
      \node[ournode] at (5.25,2.4) {};%
      \node at (5.3,2.4) [anchor=west] {Ours};%
      
      \draw[very thick] (0.82,5.1) -- ++(0,-0.4);
      \draw[very thick] (0.82,-.1) -- ++(0,0.4);
      \node at (0.82,5.65) {\scriptsize $150$};
      \node at (0.82,5.3) {\scriptsize fps};
      \draw[very thick] (1.22,5.1) -- ++(0,-0.4);
      \draw[very thick] (1.22,-.1) -- ++(0,0.4);
      \node at (1.22,5.65) {\scriptsize $60$};
      \node at (1.22,5.3) {\scriptsize fps};
      \draw[very thick] (1.52,5.1) -- ++(0,-0.4);
      \draw[very thick] (1.52,-.1) -- ++(0,0.4);
      \node at (1.52,5.65) {\scriptsize $30$};
      \node at (1.52,5.3) {\scriptsize fps};
      
      \draw[brown,thick] (0.85,3.21)--(1.15,1.85)--(1.48,1.50)--(1.83,1.17)--(2.11,1.14);
      
      \node[cpunode] at (4.63,{3.558-2}) {};%
      \node[cpunode] at (4.72,{3.571-2}) {};%
      \node[cpunode] at (4.36,{3.055-2}) {};%
      \node[cpunode] at (4.81,{5.961-2}) {};%
      \node[gpunode] at (3.80,{6.318-2}) {};%
      \node[cpunode] at (2.15,{5.178-2}) {};%
      \node[cpunode] at (3.52,{4.520-2}) {};%
      \node[cpunode] at (1.85,{6.309-2}) {};%
      \node[gpunode] at (1.26,{5.45 -2}) {};%
      \node[gpunode] at (1.51,{5.87 -2}) {};%
      \node[ournode] at (0.85,{5.21 -2}) {};%
      \node[ournode] at (1.15,{3.85 -2}) {};%
      \node[ournode] at (1.48,{3.50 -2}) {};%
      \node[ournode] at (1.83,{3.17 -2}) {};%
      \node[ournode] at (2.11,{3.14 -2}) {};%
      \node[l]at({4.63-.1},{3.558-2})[anchor=east]{EpicFlow};%
      \node[l]at({4.72+.1},{3.571-2})[anchor=west]{DeepFlow};%
      \node[l]at({4.36+.1},{3.055-2})[anchor=west]{FlowField};%
      \node[l]at({4.81+.1},{5.961-2})[anchor=west]{LDOF};%
      \node[l]at({3.80+.1},{6.318-2})[anchor=west]{LDOF (GPU)};%
      \node[l]at({2.15+.1},{5.178-2})[anchor=west]{PCA-Flow};%
      \node[l]at({3.52+.1},{4.520-2})[anchor=west]{PCA-Layers};%
      \node[l]at({1.85+.1},{6.309-2})[anchor=west]{DIS-Fast};%
      \node[l]at({1.26+.1},{5.45 -2+.04})[anchor=west]{FlowNetS};%
      \node[l]at({1.51+.1},{5.87 -2+.02})[anchor=west]{FlowNetC};%
      \node[l]at({0.85-.1},{5.21-2})[anchor=east]{FN2-s};%
      \node[l]at({1.15-.1},{3.85-2})[anchor=east]{FN2-ss};%
      \node[l]at({1.48-.11},{3.50-2.0})[anchor=north east]{FN2-css-ft-sd};%
      \node[l]at({1.83-.05},{3.17-2.1})[anchor=north east]{FN2-CSS-ft-sd};%
      \node[l]at({2.11+.1},{3.14-2})[anchor=west]{FlowNet2};%
    \end{tikzpicture}%
  \end{center}%
  \caption{Runtime vs. endpoint error comparison to the fastest existing methods with available code. The FlowNet2 family outperforms other methods by a large margin. The behaviour for the KITTI dataset is the same; see supplemental material.}%
  \label{fig:benchmark_plots}%
\end{figure}%

\begin{table*}[t]
  \begin{center}%
        \resizebox{\textwidth}{!}{%
      \begin{tabular}{|c|l||cc|cc||cc|ccc||cc||cc|}%
        \hline 
               & Method
               & \multicolumn{2}{c|}{Sintel \textit{clean}}%
               & \multicolumn{2}{c||}{Sintel \textit{final}}%
               & \multicolumn{2}{c|}{KITTI 2012}%
               & \multicolumn{3}{c||}{KITTI 2015}%
               & \multicolumn{2}{c||}{Middlebury}%
               & \multicolumn{2}{c|}{Runtime}%
               \\%
               
               &
               & \multicolumn{2}{c|}{AEE}%
               & \multicolumn{2}{c||}{AEE}%
               & \multicolumn{2}{c|}{AEE}%
               & AEE
               & Fl-all
               & Fl-all
               & \multicolumn{2}{c||}{AEE}%
               & \multicolumn{2}{c|}{ms per frame}%
               \\%
               
               &
               & \textit{train} & \textit{test}%
               & \textit{train} & \textit{test}%
               & \textit{train} & \textit{test}%
               & \textit{train}&\textit{train}&\textit{test}%
               & \textit{train} & \textit{test}%
               & CPU & GPU%
               \\%
               
        \hline \hline 

        \multirow{6}{*}{\rotatebox[origin=c]{90}{Accurate}} &
        EpicFlow$^\dagger$ \cite{epicflow}%
        & $2.27$ & $4.12$ 
        & $3.56$ & $\pz6.29$ 
        & $\pz\textbf{3.09}$ & $\pz\textbf{3.8}$ 
        & $\pz9.27$ & $27.18\%$ & $\textbf{27.10}\%$ 
        & $0.31$ & $0.39$ 
        & $42,\!600$ & -- 
        \\%

        &
        DeepFlow$^\dagger$ \cite{deepflow}%
        & $2.66$ & $5.38$  
        & $3.57$ & $\pz7.21$ 
        & $\pz4.48$ & $\pz5.8$ 
        & $10.63$ & $26.52\%$ & $29.18\%$ 
        & $\textbf{0.25}$ & $0.42$ 
        & $51,\!940$ & -- 
        \\%

        &
        FlowFields \cite{flowfields}%
        & $\textbf{1.86}$ & $\textbf{3.75}$  
        & $\textbf{3.06}$ & $\pz\textbf{5.81}$ 
        & $\pz3.33$ & $\pz3.5$ 
        & $\pz\textbf{8.33}$ & $\textbf{24.43\%}$ & -- 
        & $0.27$ & $\textbf{0.33}$ 
        & $22,\!810$ & -- 
        \\%

        &
        LDOF (CPU) \cite{ldof}%
        & $4.64$ & $7.56$  
        & $5.96$ & $\pz9.12$
        & $10.94$ & $12.4$ 
        & $18.19$ & $38.11\%$ & -- 
        & $0.44$ & $0.56$ 
        & $64,\!900$ & -- 
        \\%

        &
        LDOF (GPU) \cite{ldof-gpu}%
        & $4.76$ & --  
        & $6.32$ & -- 
        & $10.43$ & -- 
        & $18.20$ & $38.05\%$ & -- 
        & $0.36$ & -- 
        & -- & $\phantom{,}6,\!270$ 
        \\%

        &
        PCA-Layers \cite{pcaflowlayers}%
        & $3.22$ & $ 5.73$  
        & $4.52$ & $\pz7.89$ 
        & $\pz5.99$ & $\pz5.2$ 
        & $12.74$ & $27.26\%$ & -- 
        & $0.66$ & -- 
        & $\pz\textbf{3,300}$ & -- 
        \\%

        \hline \hline 

        \multirow{5}{*}{\rotatebox[origin=c]{90}{Fast}} &
        EPPM \cite{eppm}%
        & -- & $\textbf{6.49}$  
        & -- & $\pz\textbf{8.38}$ 
        & -- & $\pz9.2$ 
        & -- & -- & -- 
        & -- & $\textbf{0.33}$ 
        & -- & $\pz\pz200$ 
        \\%

        &
        PCA-Flow \cite{pcaflowlayers}%
        & $\textbf{4.04}$ & $6.83$  
        & $\textbf{5.18}$ & $\pz8.65$ 
        & $\pz\textbf{5.48}$ & $\pz\textbf{6.2}$ 
        & $\textbf{14.01}$ & $\textbf{39.59\%}$ & -- 
        & $\textbf{0.70}$ & -- 
        & $\pz\pz\phantom{,}140$ & -- 
        \\%

        &
        DIS-Fast \cite{dis-fast}%
        & $5.61$ & $9.35$  
        & $6.31$ & $10.13$ 
        & $11.01$ & $14.4$ 
        & $21.20$ & $53.73\%$ & -- 
        & $0.92$ & -- 
        & $\pz\pz\phantom{,}\pz70$ & -- 
        \\%

        &
        FlowNetS \cite{DFIB15}%
        & $4.50$ & $\phantom{^\ddagger}6.96^\ddagger$  
        & $5.45$ & $\pz\phantom{^\ddagger}7.52^\ddagger$  
        & $\pz8.26$ & -- 
        & -- & -- & --
        & $1.09$ & -- 
        & -- & $\pz\pz\pz\textbf{18}$ 
        \\%
        

        &
        FlowNetC \cite{DFIB15}%
        & $4.31$ & $\phantom{^\ddagger}6.85^\ddagger$  
        & $5.87$ & $\pz\phantom{^\ddagger}8.51^\ddagger$ 
        & $\pz9.35$ & -- 
        & -- & -- & --
        & $1.15$ & -- 
        & -- & $\pz\pz\pz32$ 
        \\%


        \hline \hline 

        \multirow{10}{*}{\rotatebox[origin=c]{90}{FlowNet 2.0}} &
        FlowNet2-s  
        & $4.55$ & --  
        & $5.21$  & -- 
        & $\pz8.89$ & -- 
        & $16.42$ & $56.81\%$ & -- 
        & $1.27$ & -- 
        & -- & $\pz\pz\pz\pz\textbf{7}$
        \\%

        &
        FlowNet2-ss 
        & $3.22$ & -- 
        & $3.85$ & -- 
        & $\pz5.45$ & -- 
        & $12.84$ & $41.03\%$ & -- 
        & $0.68$ & -- 
        & -- & $\pz\pz\pz14$
        \\%

        &
        FlowNet2-css 
        & $2.51$ & -- 
        & $3.54$ & -- 
        & $\pz4.49$ & -- 
        & $11.01$ & $35.19\%$ & -- 
        & $0.54$ & -- 
        & -- & $\pz\pz\pz31$
        \\%

        &
        FlowNet2-css-ft-sd
        & $2.50$ & -- 
        & $3.50$ & -- 
        & $\pz4.71$ & -- 
        & $11.18$ & $34.10\%$ & -- 
        & $0.43$ & -- 
        & -- & $\pz\pz\pz31$
        \\%

        &
        FlowNet2-CSS 
        & $2.10$ & -- 
        & $3.23$ & -- 
        & $\pz\textbf{3.55}$ & -- 
        & $\textbf{\pz8.94}$ & $29.77\%$ & -- 
        & $0.44$ & -- 
        & -- & $\pz\pz\pz69$
        \\%

        &
        FlowNet2-CSS-ft-sd 
        & $2.08$  & -- 
        & $3.17$  & -- 
        & $\pz4.05$ & -- 
        & $10.07$ & $30.73\%$ & -- 
        & $0.38$ & -- 
        & -- & $\pz\pz\pz69$
        \\%


        &
        FlowNet2     
        \cellcolor{gray!15}%
        & \cellcolor{gray!15}$\textbf{2.02}$ & \cellcolor{gray!15}$\textbf{3.96}$ 
        & \cellcolor{gray!15}$\textbf{3.14}$ & \cellcolor{gray!15}$6.02$ 
        & \cellcolor{gray!15}$\pz4.09$  & \cellcolor{gray!15}--  
        & \cellcolor{gray!15}$10.06$ & \cellcolor{gray!15}$30.37\%$ & \cellcolor{gray!15}-- 
        & \cellcolor{gray!15}$\textbf{0.35}$ & \cellcolor{gray!15}$\textbf{0.52}$ 
        & \cellcolor{gray!15}-- & \cellcolor{gray!15}$\pz\pz123$
        \\%

        &
        FlowNet2-ft-sintel  
        & ($1.45$) & $4.16$ 
        & ($2.01$) & $\textbf{5.74}$ 
        & $\pz3.61$ & -- 
        & $\pz9.84$ & $\textbf{28.20\%}$ & -- 
        & $0.35$ & -- 
        & -- & $\pz\pz123$
        \\%

        &
        FlowNet2-ft-kitti
        & $3.43$ & -- 
        & $4.66$ & -- 
        & $\pz$($1.28$) & \textbf{1.8} 
        & $\pz$($2.30$) & $\pz$($8.61\%$) & $\textbf{11.48\%}$ 
        & $0.56$ & -- 
        & -- & $\pz\pz123$
        \\%

        \hline 
      \end{tabular}%
    }%

  \end{center}%
  \caption{Performance comparison on public benchmarks. AEE: Average Endpoint Error; Fl-all: Ratio of pixels where flow estimate is wrong by both $\geq3$ pixels and $\geq5$\%. The best number for each category is highlighted in bold. See text for details. $^\dagger$\textit{train} numbers for these methods use slower but better "improved" option. $^\ddagger$For these results we report the fine-tuned numbers (FlowNetS-ft and FlowNetC-ft).
  }%
  \label{tab:benchmark_results}%
\end{table*}%

We evaluated all methods\footnote{An exception is EPPM for which we could not provide the required Windows environment and use the results from \cite{eppm}.} on a system with an Intel Xeon E5 with 2.40GHz and an Nvidia GTX 1080. Where applicable, dataset-specific parameters were used, that yield best performance. Endpoint errors and runtimes are given in Table~\ref{tab:benchmark_results}. 

\textbf{Sintel:} On Sintel, FlowNet2 consistently outperforms DeepFlow~\cite{deepflow} and EpicFlow~\cite{epicflow} and is on par with FlowFields. All methods with comparable runtimes have clearly inferior accuracy. We fine-tuned FlowNet2 on a mixture of Sintel \textit{clean}+\textit{final} training data (\FN{-ft-sintel}). On the benchmark, in case of \textit{clean} data this slightly degraded the result, while on \textit{final} data \FN{-ft-sintel} is on par with the currently published state-of-the art method DeepDiscreteFlow~\cite{deepdiscreteflow}. 

\textbf{KITTI:} On KITTI, the results of \FN{CSS} are comparable to EpicFlow~\cite{epicflow} and FlowFields~\cite{flowfields}. Fine-tuning on small displacement data degrades the result. This is probably due to KITTI containing very large displacements in general. Fine-tuning on a combination of the KITTI2012 and KITTI2015 training sets reduces the error roughly by a factor of $3$ (\FN{ft-kitti}). Among non-stereo methods we obtain the best EPE on KITTI2012 and the first rank on the KITTI2015 benchmark. This shows how well and elegantly the learning approach can integrate the prior of the driving scenario. 

\textbf{Middlebury:} On the Middlebury training set FlowNet2 performs comparable to traditional methods. The results on the Middlebury test set are unexpectedly a lot worse. Still, there is a large improvement compared to FlowNetS~\cite{DFIB15}. 

Endpoint error vs. runtime evaluations for Sintel are provided in Figure~\ref{fig:benchmark_plots}. One can observe that the FlowNet2 family outperforms the best and fastest existing methods by large margins. Depending on the type of application, a FlowNet2 variant between 8 to 140 frames per second can be used. 

\subsection{Qualitative Results}

\begin{figure*}[t]
  \begin{center}%
    \newcommand{\galleryRowSintel}[6]{\vspace*{-0.075cm} \includegraphics[width=0.2\textwidth]{figures/gallery/#1} & \includegraphics[width=0.2\textwidth]{figures/gallery/#2} &  \includegraphics[width=0.2\textwidth]{figures/gallery/#3} &  \includegraphics[width=0.2\textwidth]{figures/gallery/#4} &  \includegraphics[width=0.2\textwidth]{figures/gallery/#5} &  \includegraphics[width=0.2\textwidth]{figures/gallery/#6} \tabularnewline 
}
  \resizebox{\linewidth}{!}{%
    \setlength{\tabcolsep}{0.7pt}%
    \begin{tabular}{cccccc}%
      \multirow{2}{*}{Image Overlay} &
      \multirow{2}{*}{Ground Truth} &
      FlowFields \cite{flowfields} &
      PCA-Flow \cite{pcaflowlayers} &
      FlowNetS \cite{DFIB15} &
      FlowNet2 \\
      &
      &
      ($22,\!810$ms)&
      ($140$ms)&
      ($18$ms)&
      ($123$ms)\\
    \galleryRowSintel{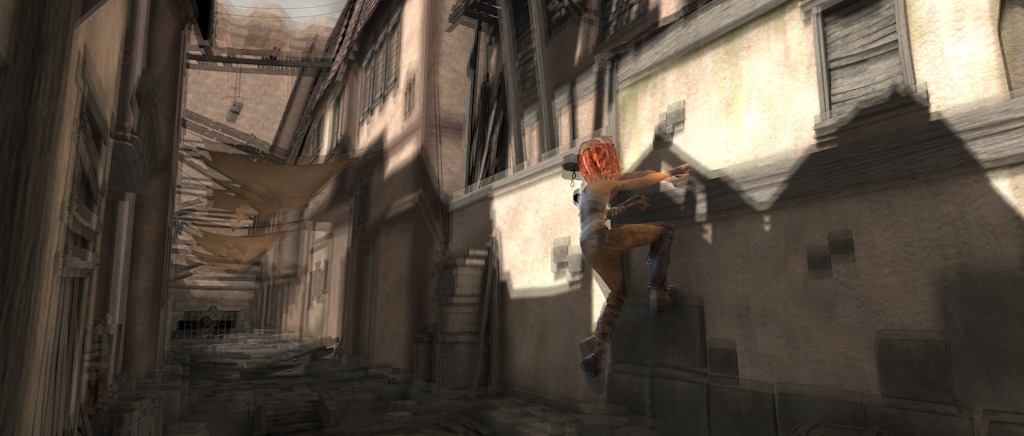}{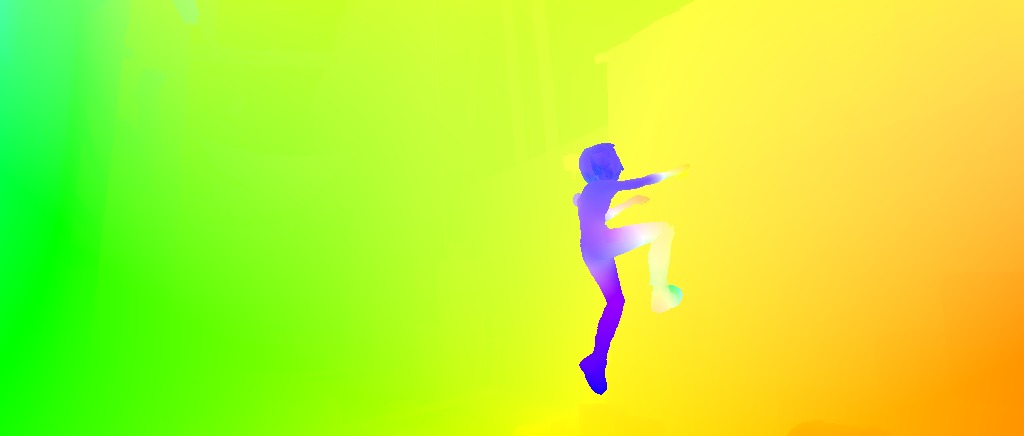}{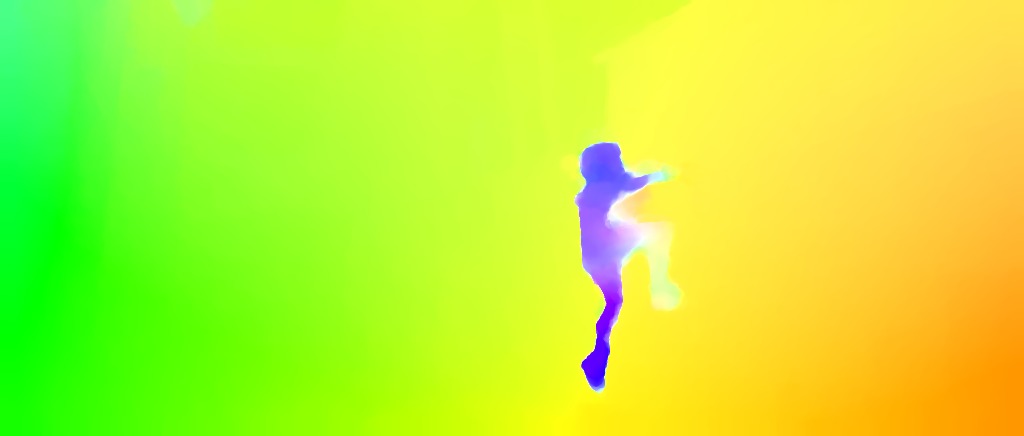}{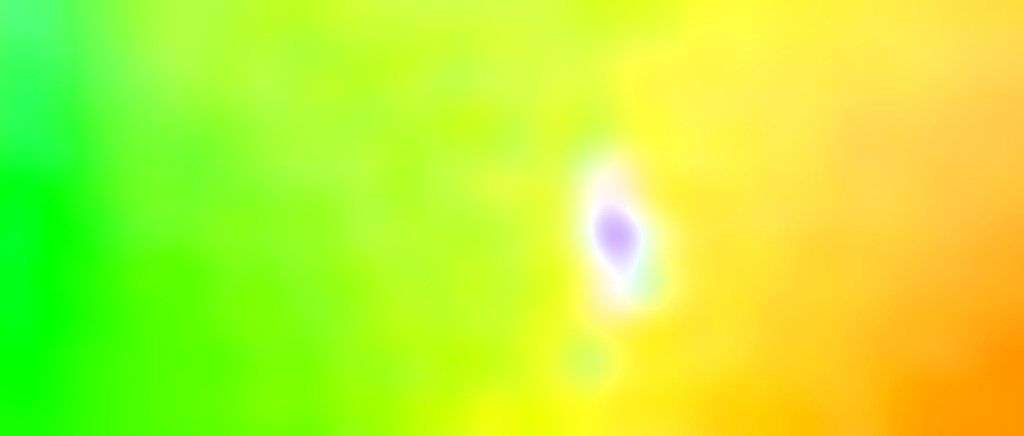}{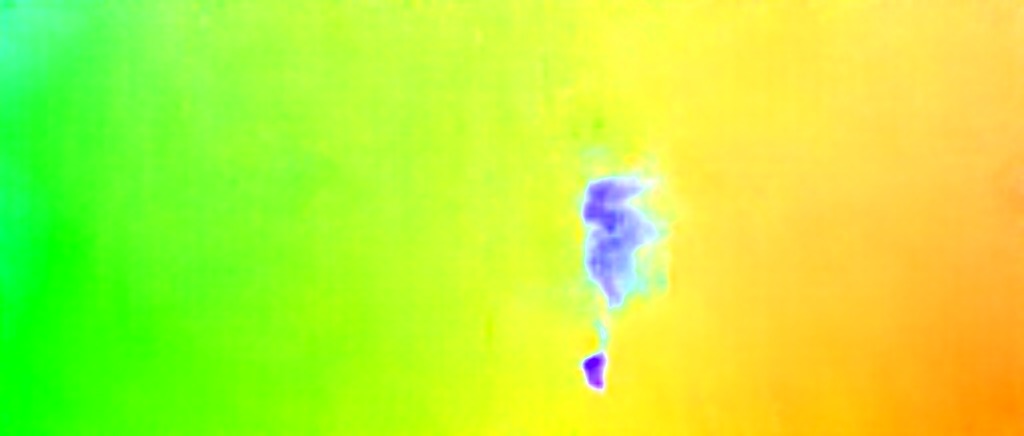}{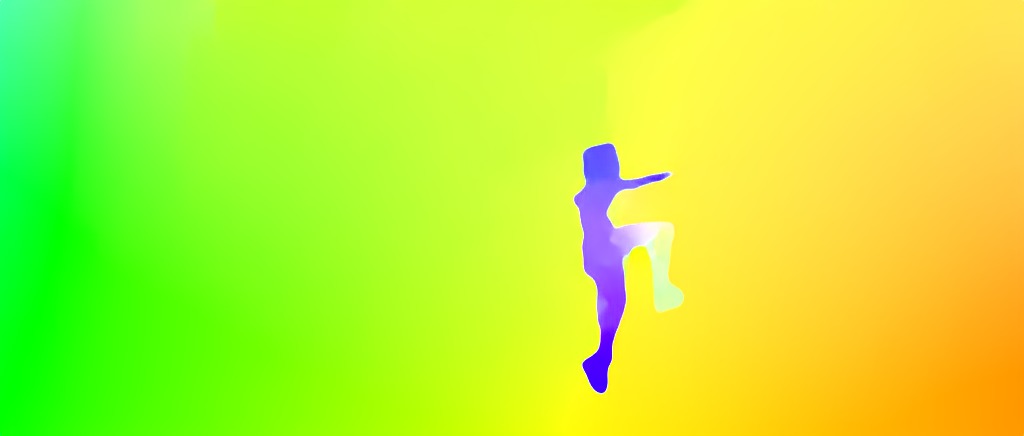}%
    \galleryRowSintel{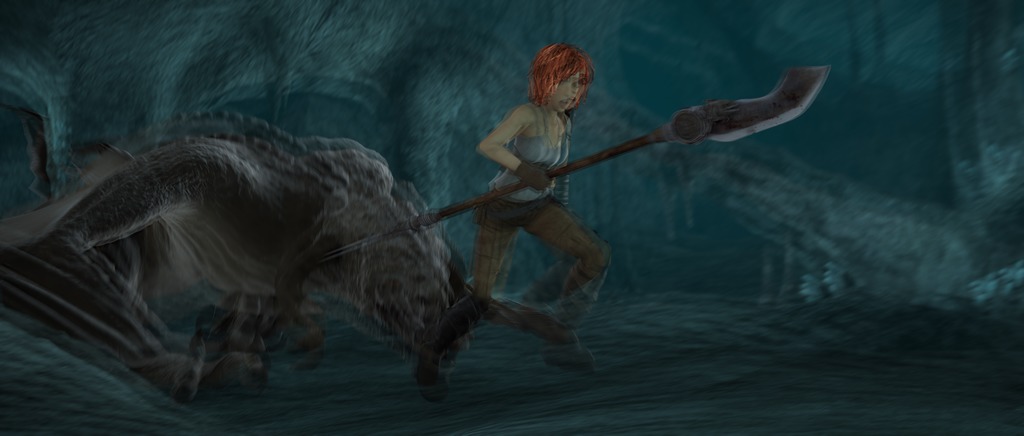}{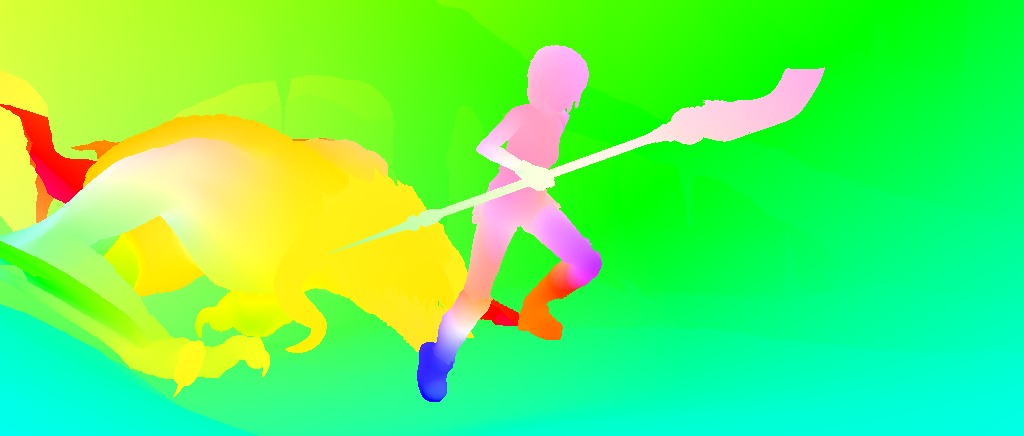}{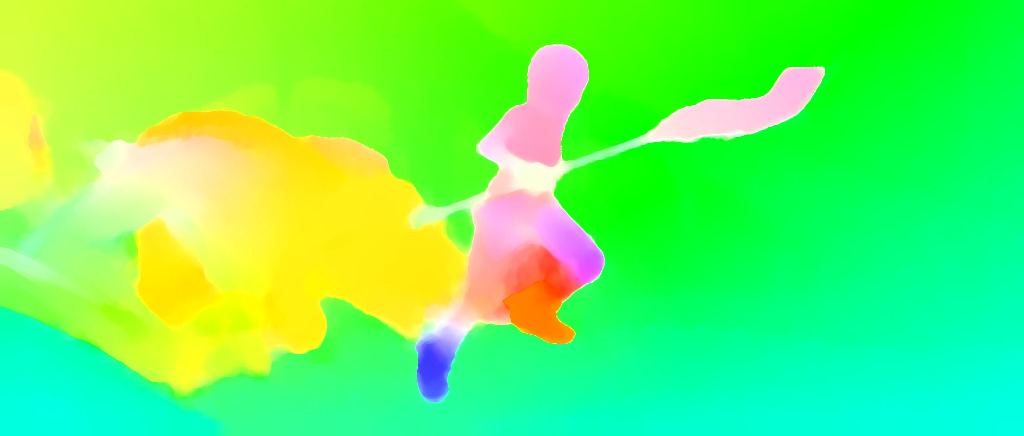}{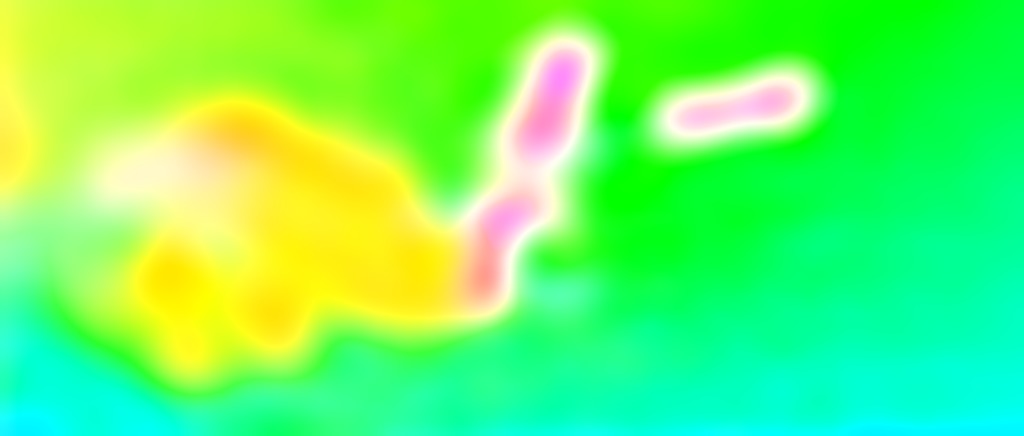}{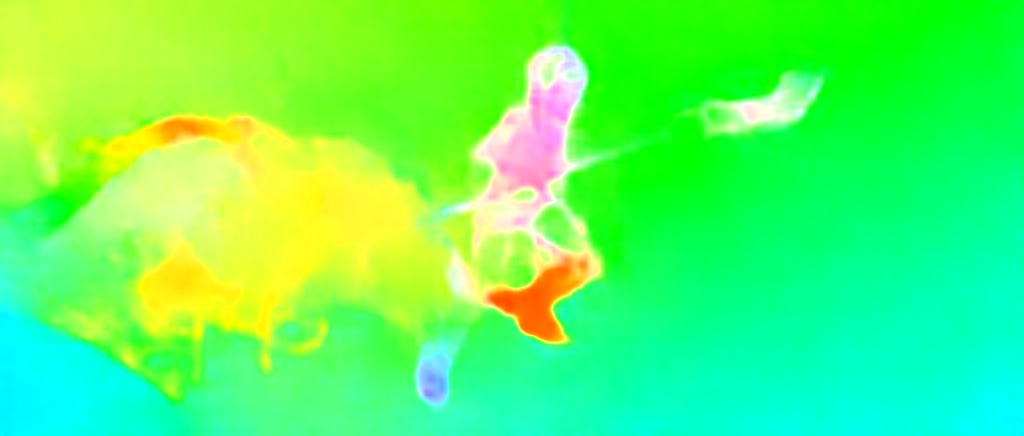}{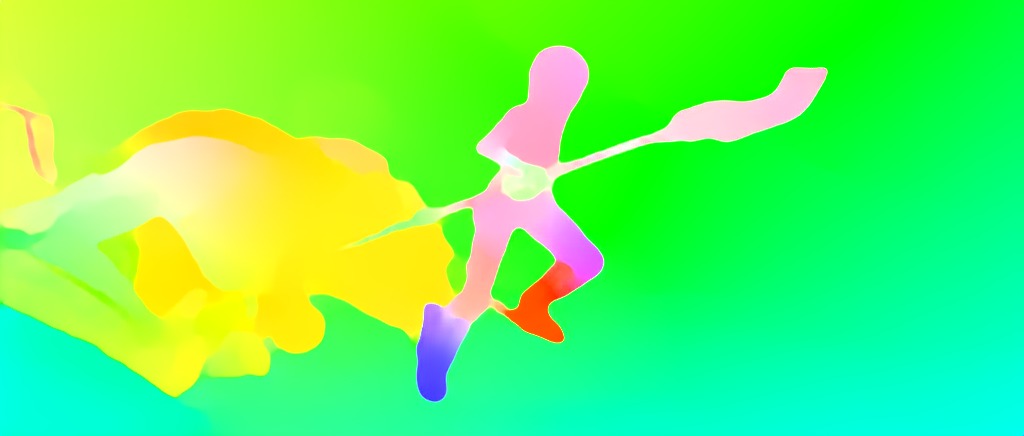}%
    \galleryRowSintel{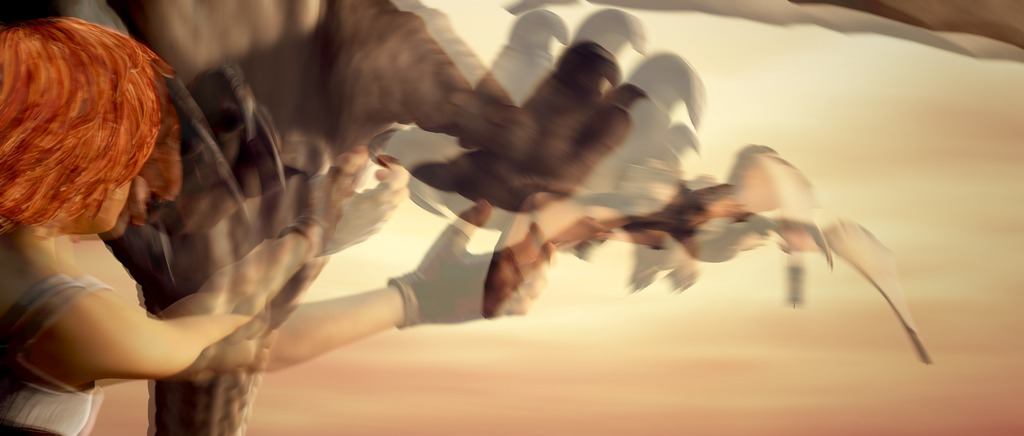}{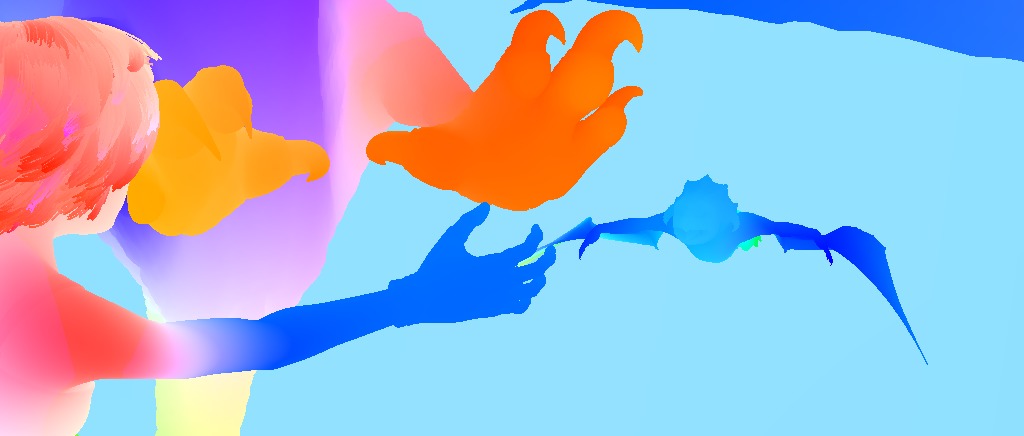}{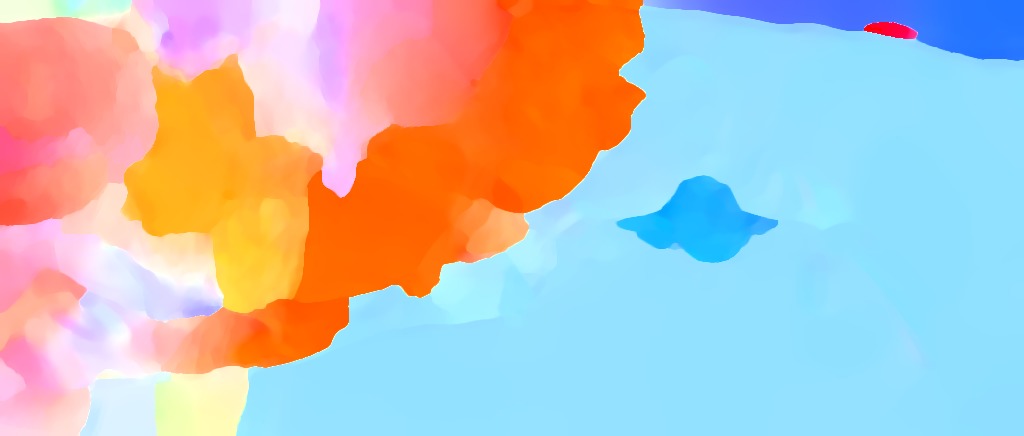}{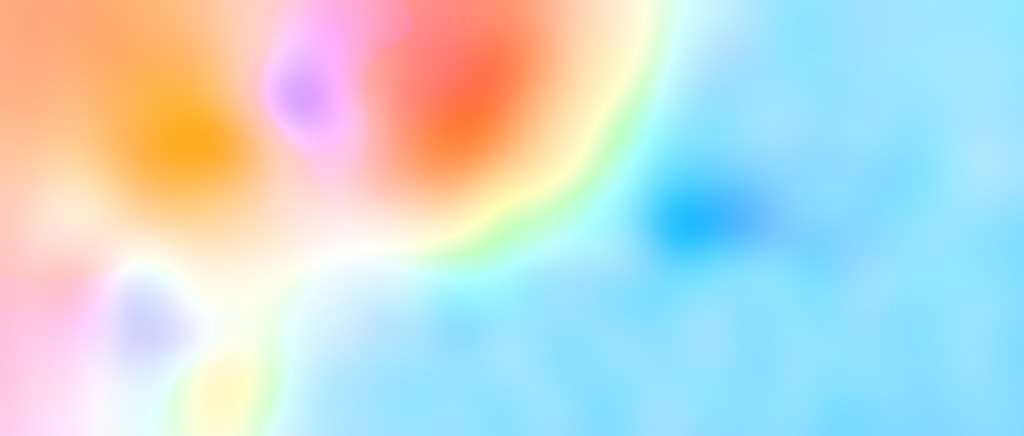}{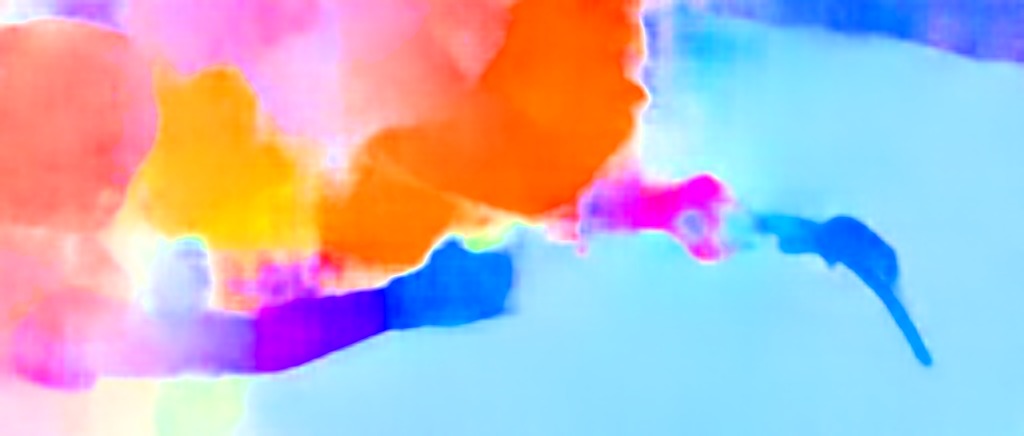}{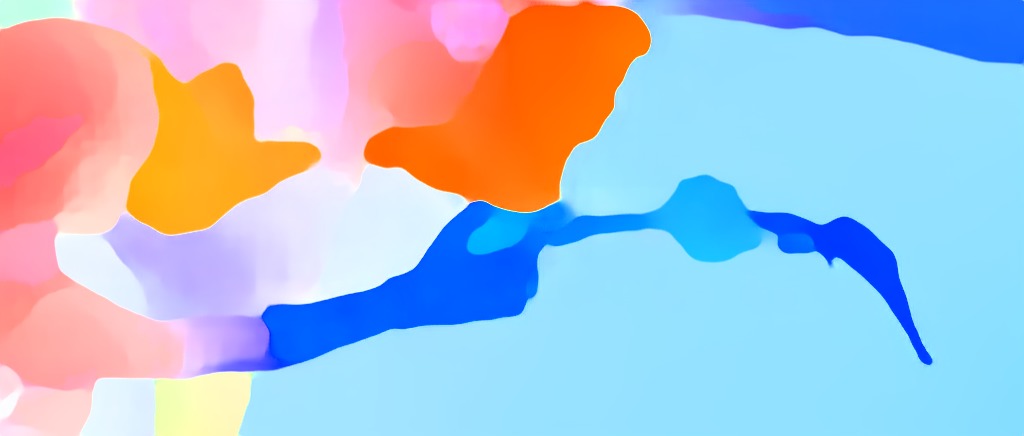}%
    \galleryRowSintel{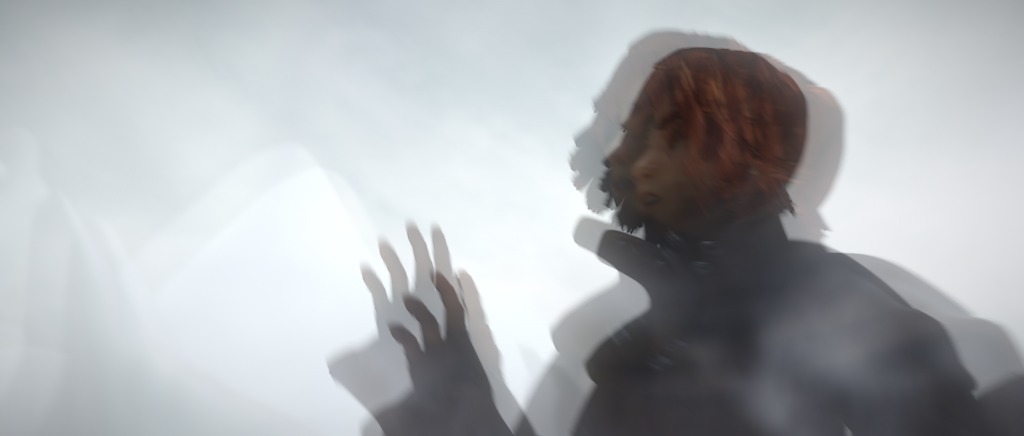}{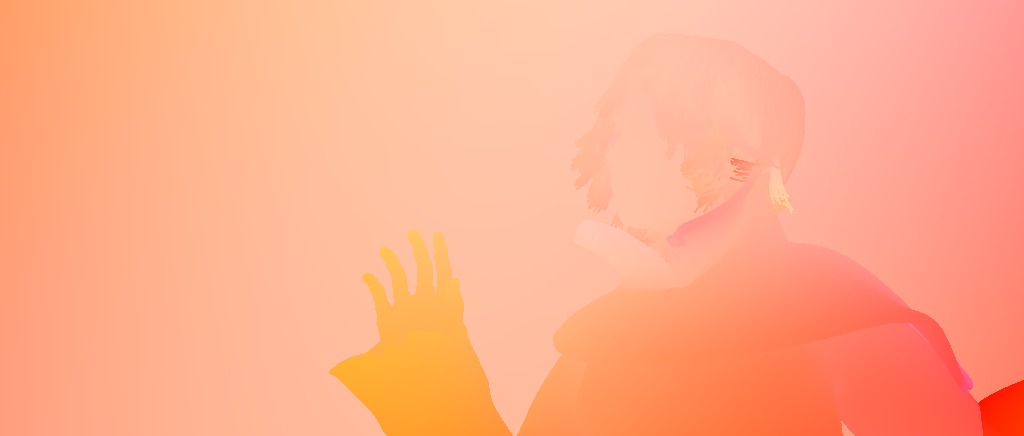}{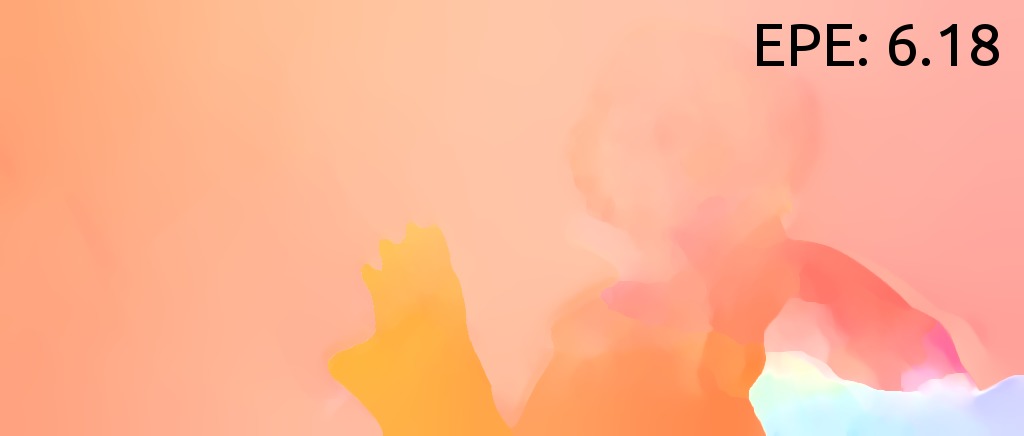}{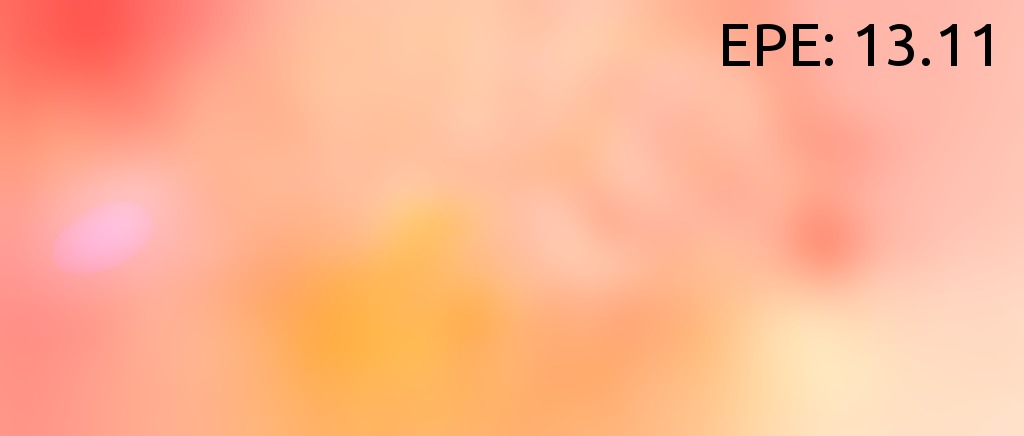}{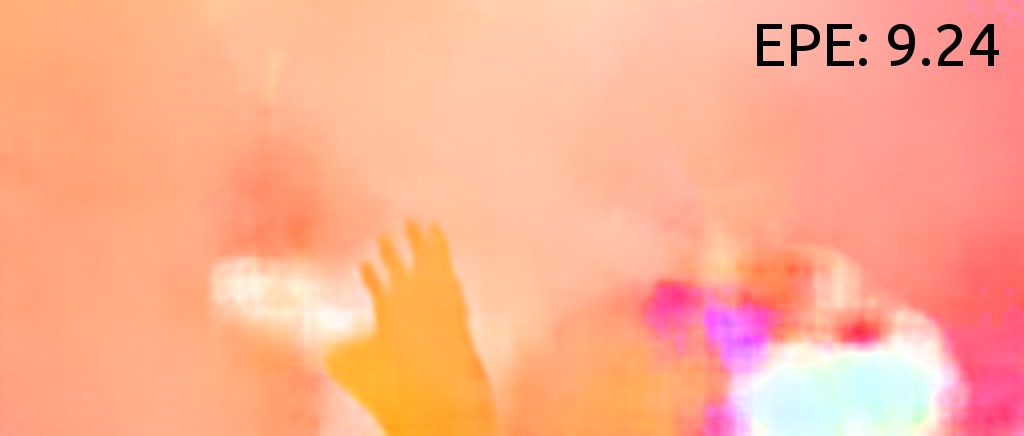}{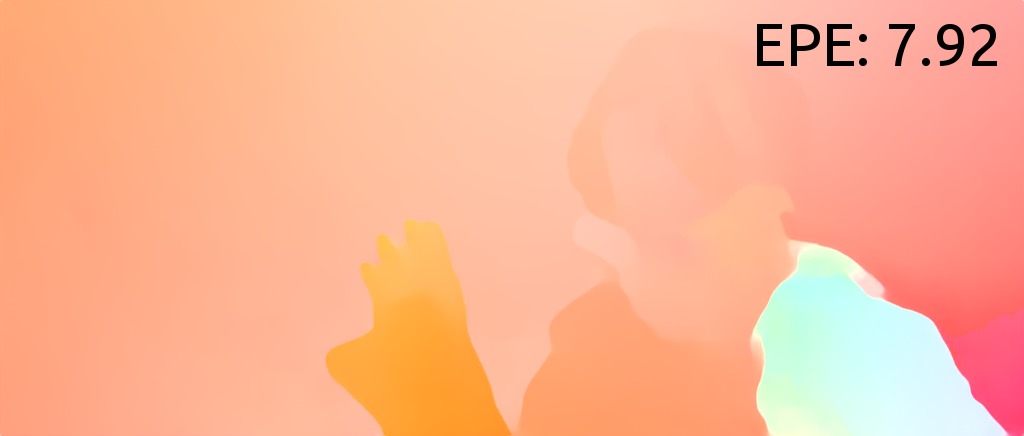}       \galleryRowSintel{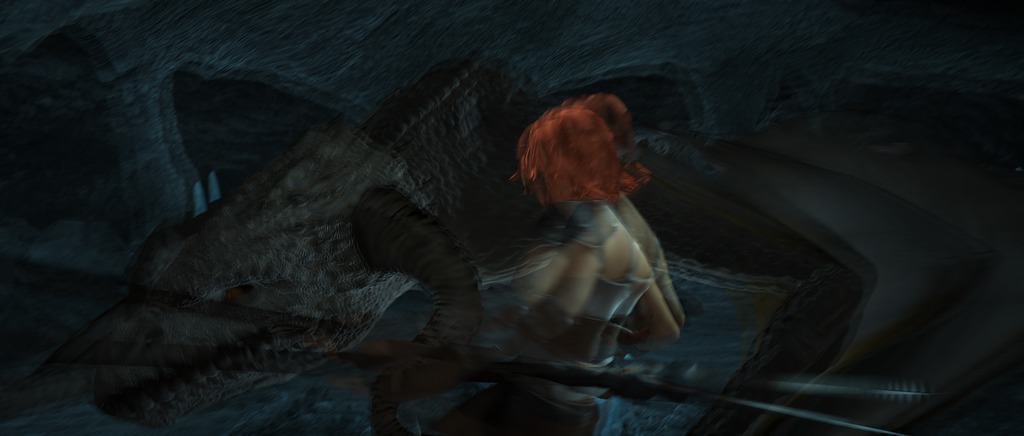}{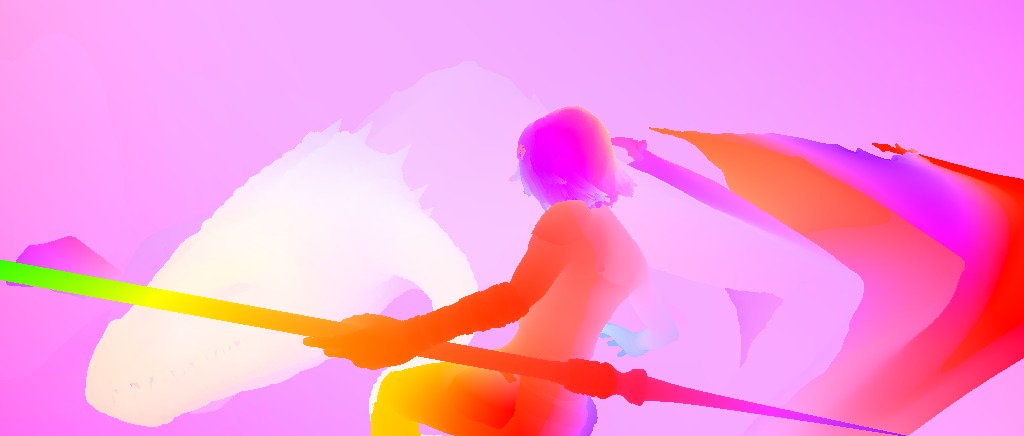}{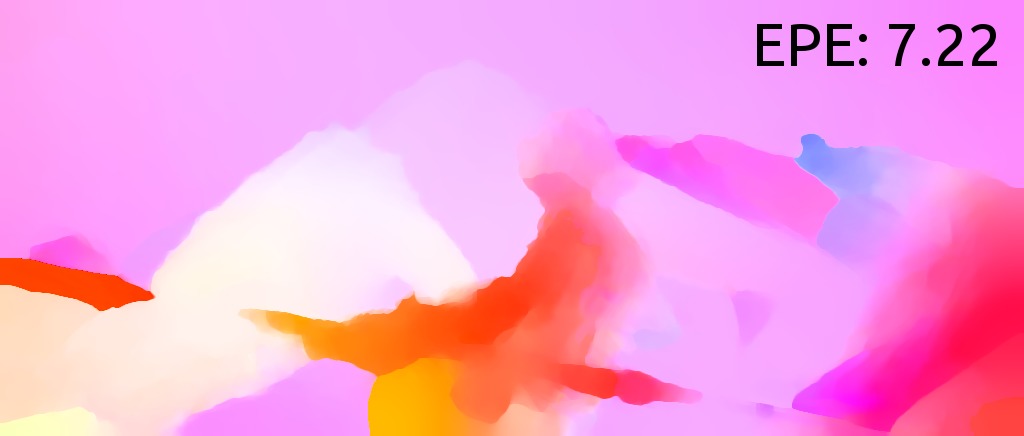}{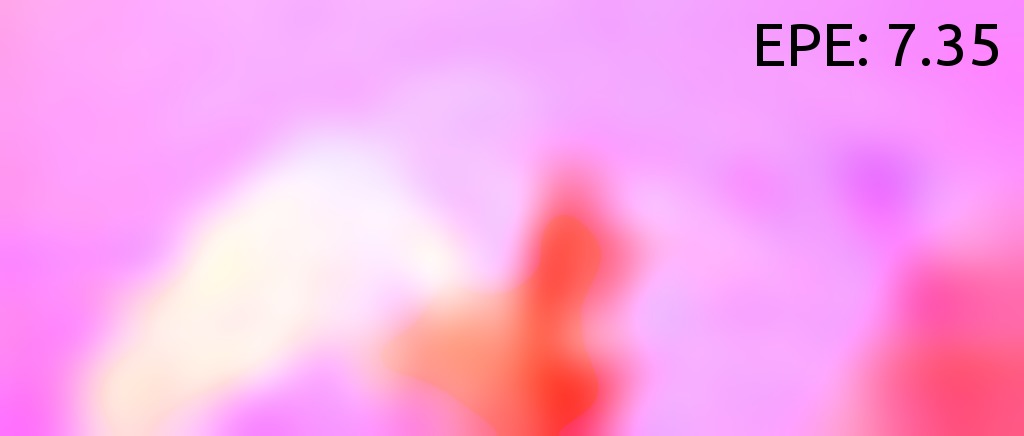}{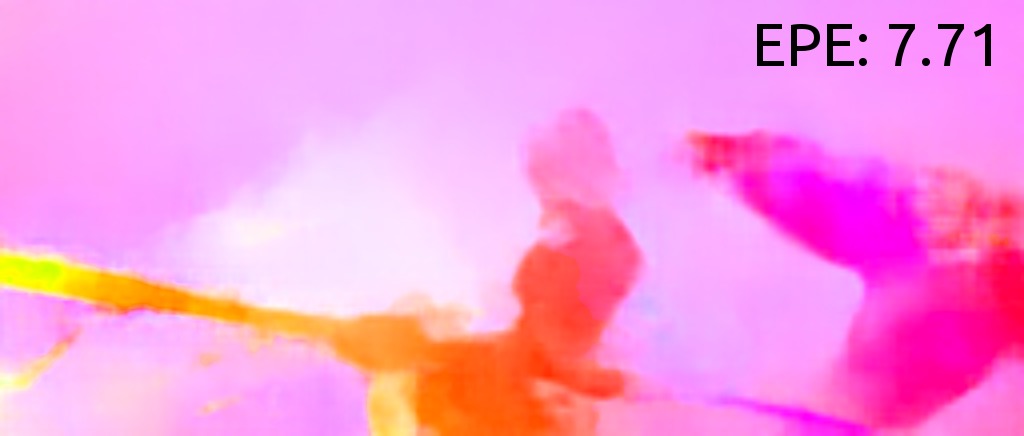}{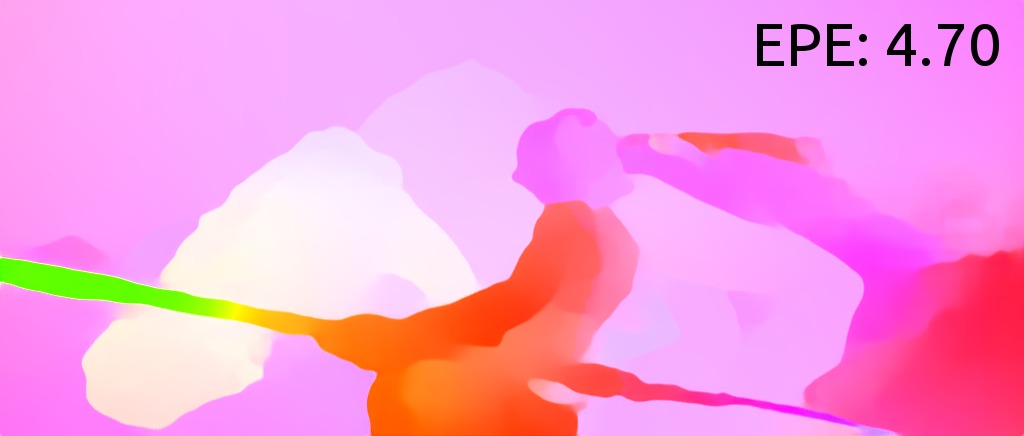}%
    \end{tabular}%
  }%
 
  \end{center}%
  \caption{Examples of flow fields from different methods estimated on Sintel. FlowNet2 performs similar to FlowFields and is able to extract fine details, while methods running at comparable speeds perform much worse (PCA-Flow and FlowNetS).}%
  \label{fig:gallery_sintel}%
\end{figure*}

Figures \ref{fig:gallery_sintel} and \ref{fig:gallery_other} show example results on Sintel and on real-world data. While the performance on Sintel is similar to FlowFields~\cite{flowfields}, we can see that on real world data FlowNet 2.0 clearly has advantages in terms of being robust to homogeneous regions (rows 2 and 5), image and compression artifacts (rows 3 and 4) and it yields smooth flow fields with sharp motion boundaries. 

\newcommand{\galleryWidth}{0.165\linewidth}%


\begin{figure*}[t]
  \begin{center}%
    \newcommand{\galleryRowOther}[7]{\vspace*{-0.075cm} \includegraphics[width=\galleryWidth]{figures/gallery/#1} & \includegraphics[width=\galleryWidth]{figures/gallery/#2} &  \includegraphics[width=\galleryWidth]{figures/gallery/#3} &  \includegraphics[width=\galleryWidth]{figures/gallery/#4} &  \includegraphics[width=\galleryWidth]{figures/gallery/#5} &  \includegraphics[width=\galleryWidth]{figures/gallery/#6} &  \includegraphics[width=\galleryWidth]{figures/gallery/#7} \tabularnewline 
}
  \resizebox{\linewidth}{!}{%
    \setlength{\tabcolsep}{0.7pt}%
    \begin{tabular}{ccccccc}%
      Image Overlay &
      FlowFields  \cite{flowfields} &
      DeepFlow \cite{deepflow} &
      LDOF (GPU) \cite{ldof-gpu} &
      PCA-Flow \cite{pcaflowlayers} &
      FlowNetS \cite{DFIB15} &
      FlowNet2 \\
\galleryRowOther{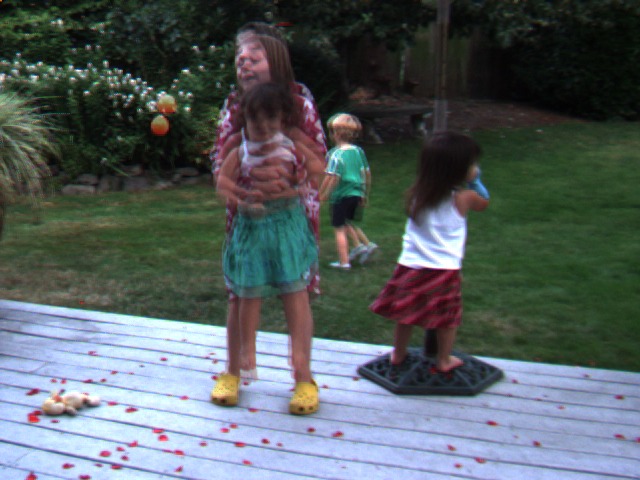}{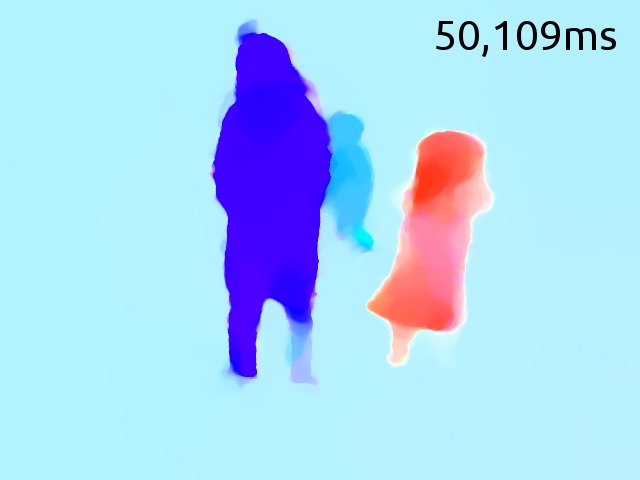}{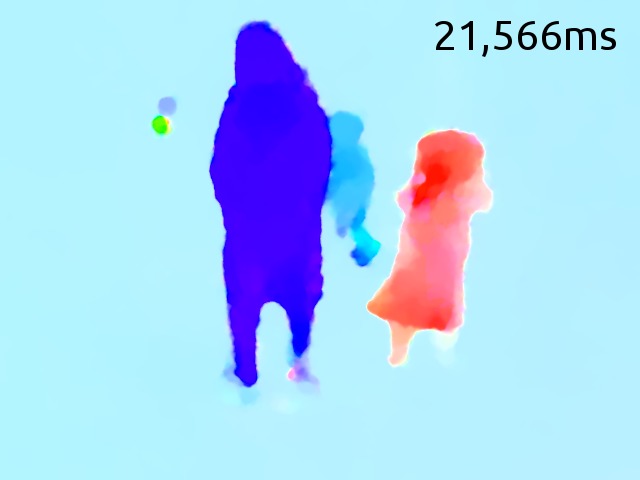}{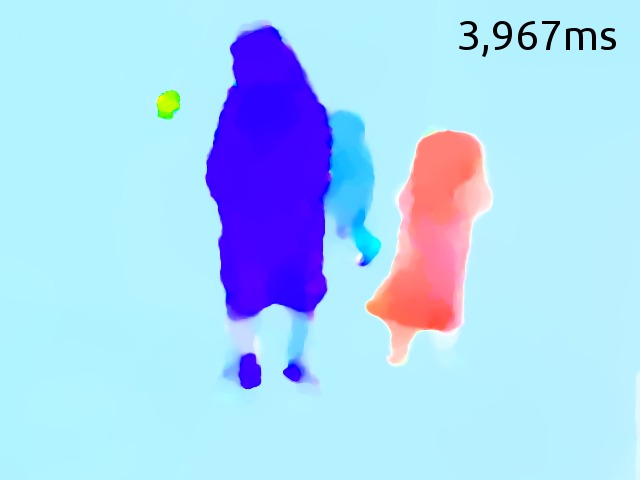}{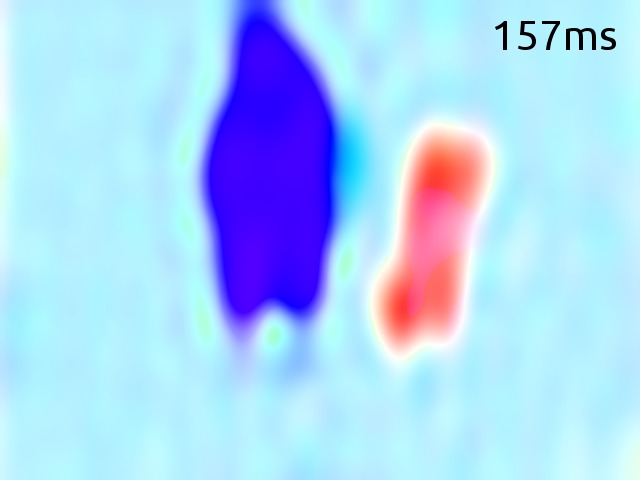}{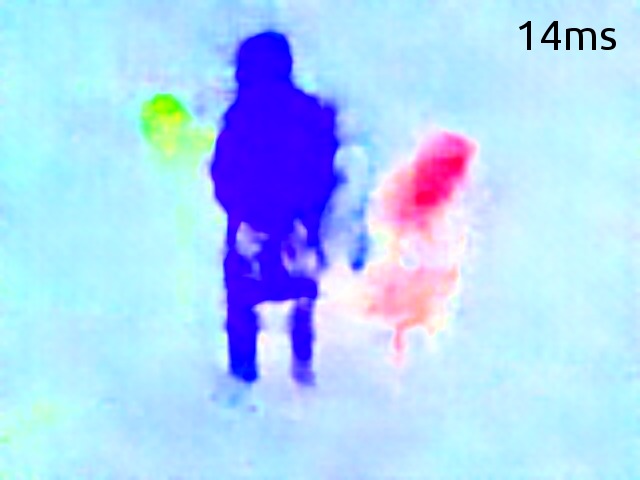}{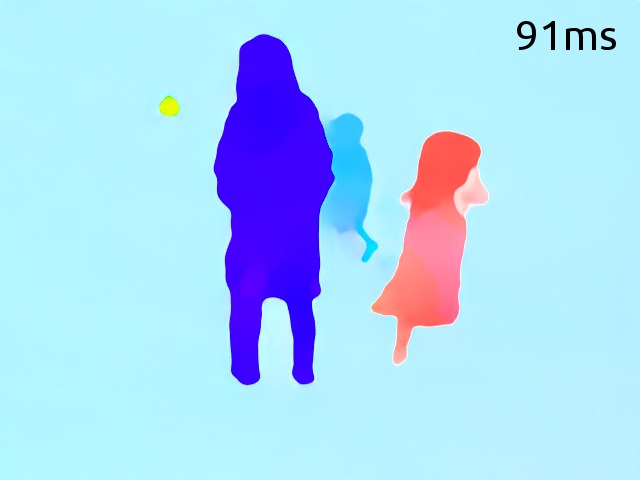}
\galleryRowOther{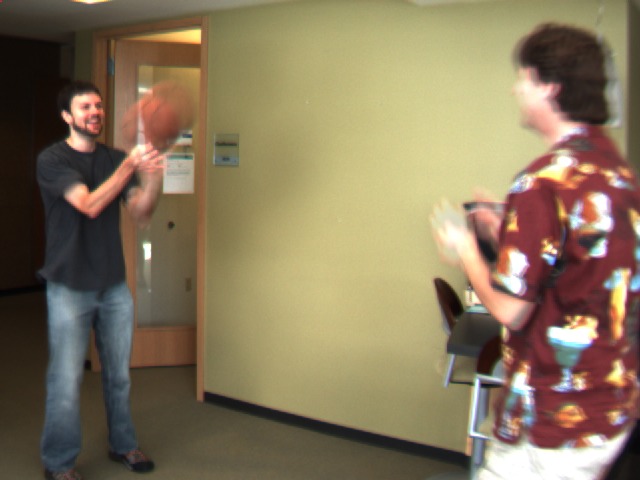}{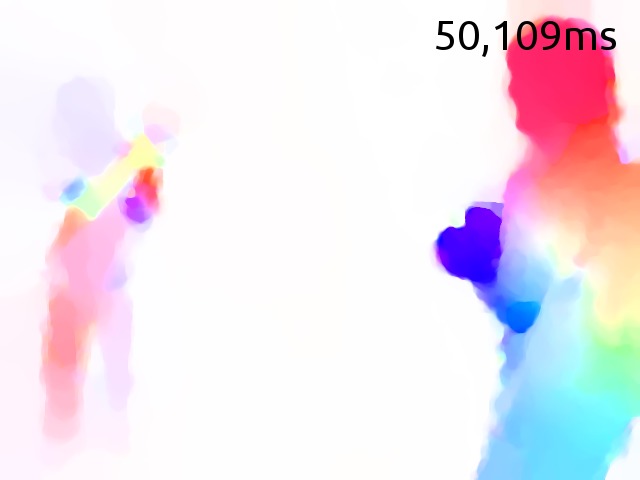}{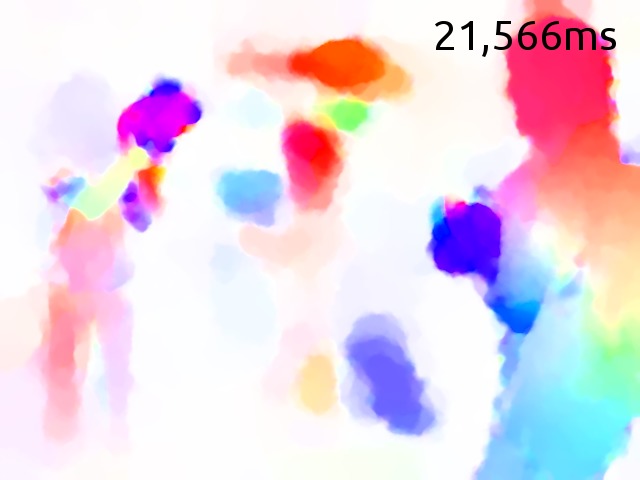}{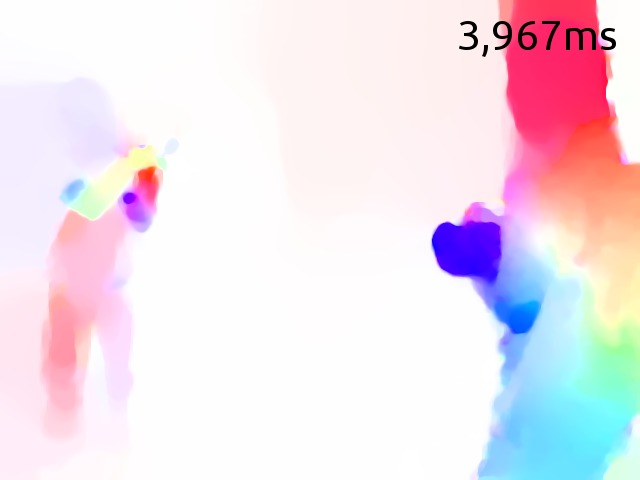}{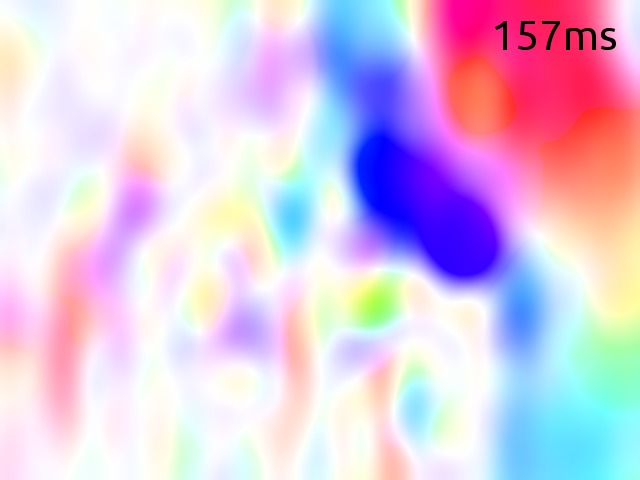}{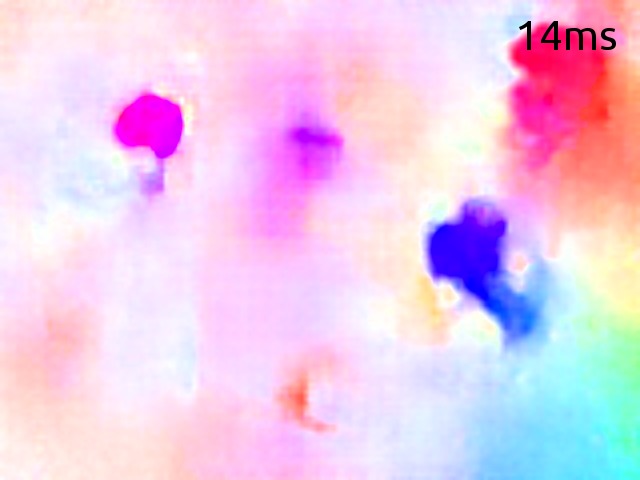}{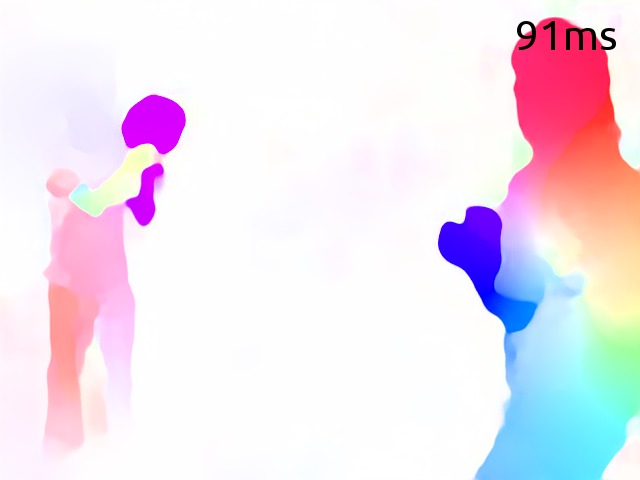}
\galleryRowOther{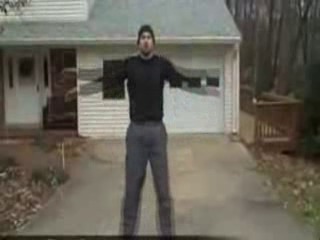}{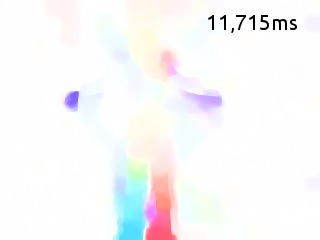}{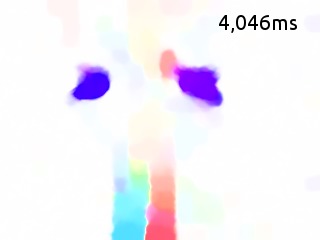}{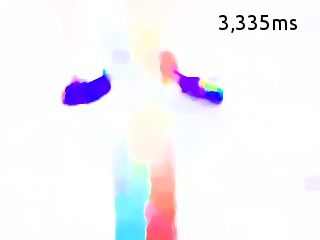}{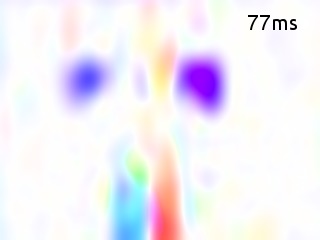}{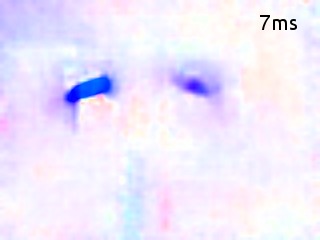}{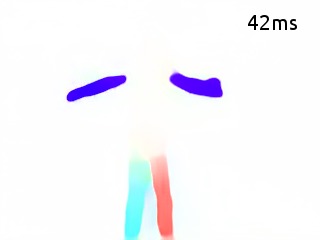}
\galleryRowOther{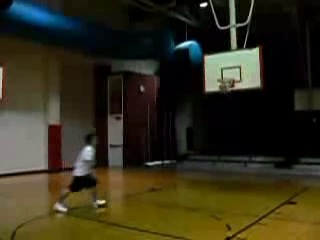}{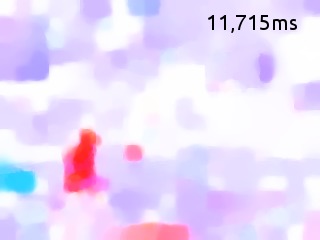}{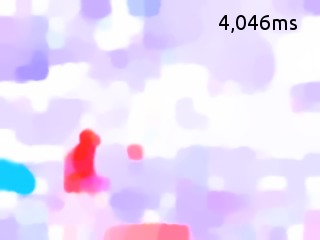}{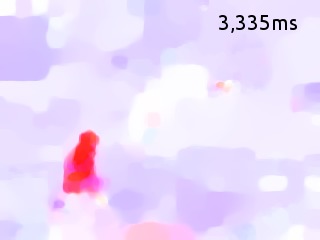}{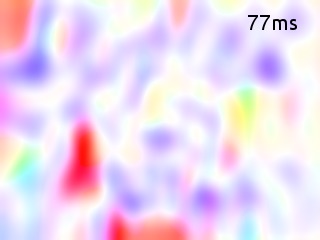}{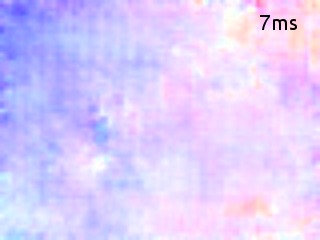}{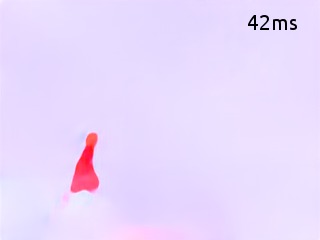}
\galleryRowOther{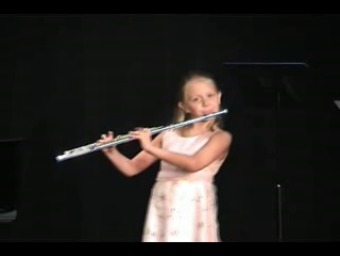}{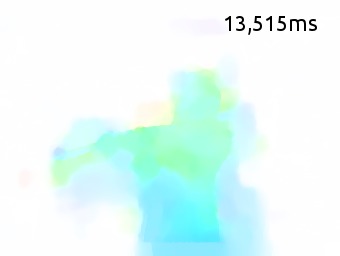}{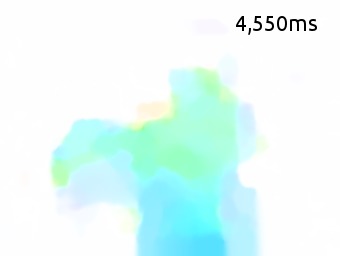}{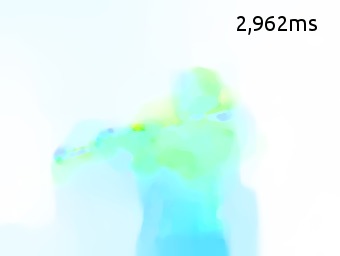}{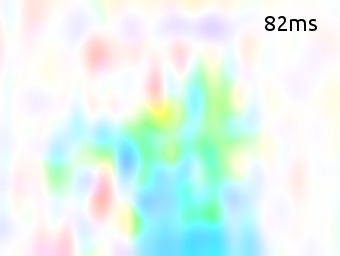}{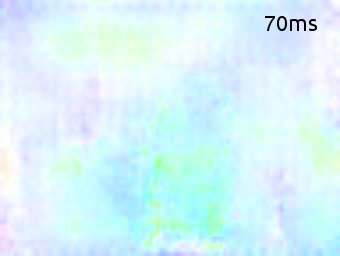}{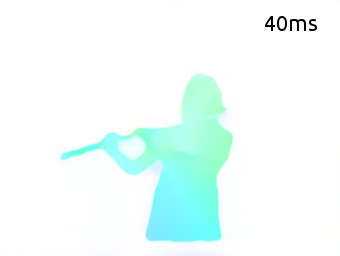}
    \end{tabular}%
  }%
 
  \end{center}%
  \caption{Examples of flow fields from different methods estimated on real-world data. The top two rows are from the Middlebury data set and the bottom three from UCF101. Note how well FlowNet2 generalizes to real-world data, i.e. it produces smooth flow fields, crisp boundaries and is robust to motion blur and compression artifacts. Given timings of methods differ due to different image resolutions.}%
  \label{fig:gallery_other}%
  \vspace*{-0.5mm}
\end{figure*}

\subsection{Performance on Motion Segmentation and Action Recognition} 

To assess performance of FlowNet 2.0 in real-world applications, we compare the performance of action recognition and motion segmentation. For both applications, good optical flow is key. Thus, a good performance on these tasks also serves as an indicator for good optical flow. 

For motion segmentation, we rely on the well-established approach of Ochs~\etal~\cite{Ochs14} to compute long term point trajectories. A motion segmentation is obtained from these using the state-of-the-art method from Keuper~\etal~\cite{KB15b}. The results are shown in Table~\ref{tab:results_moseg_actreg}. The original model in Ochs~\etal~\cite{KB15b} was built on Large Displacement Optical Flow~\cite{ldof}. We included also other popular optical flow methods in the comparison. The old FlowNet~\cite{DFIB15} was not useful for motion segmentation. In contrast, the FlowNet2 is as reliable as other state-of-the-art methods while being orders of magnitude faster. 



Optical flow is also a crucial feature for action recognition. To assess the performance, we trained the temporal stream of the two-stream approach from Simonyan~\etal~\cite{twostream} with different optical flow inputs. Table~\ref{tab:results_moseg_actreg} shows that FlowNetS~\cite{DFIB15} did not provide useful results, while the flow from FlowNet 2.0 yields comparable results to state-of-the art methods. 

\begin{table}[t]
    \begin{center}
            \resizebox{0.91\linewidth}{!}{%
        \begin{tabular}{|l||c|c||c|}%
            \hline%
                                   & \multicolumn{2}{c||}{Motion Seg.}       & Action Recog. \\%
                                   \cline{2-4} 
                                   & F-Measure & Extracted             & Accuracy \\%
                                   &           & Objects               &          \\%
            \hline \hline 
            LDOF-CPU \cite{ldof} & 
            $79.51\%$ &
            $28/65$ & 
            $\phantom{^\dagger}79.91\%^\dagger$ \\%
            
            DeepFlow \cite{deepflow} & 
            $\textbf{80.18}\%$ &
            $29/65$ & 
            \textbf{81.89}\% \\%
            
            EpicFlow \cite{epicflow} & 
            $78.36\%$ & 
            $27/65$ & 
            $78.90\%$ \\
            
            FlowFields \cite{flowfields}& 
            $79.70\%$ & 
            $\textbf{30/65}$ & 
            -- \\%
            
            \hline 
            
            FlowNetS~\cite{DFIB15} & 
            $\phantom{^\ddagger}56.87\%^\ddagger$ & 
            $\phantom{^\ddagger}\pz3/62^\ddagger$ & 
            $55.27\%$ \\%
            
            \hline
            

            FlowNet2-css-ft-sd & 
            $77.88\%$ & 
            $26/65$ & 
            -- \\%
            
            FlowNet2-CSS-ft-sd & 
            $79.52\%$ &
            $30/65$ & 
            $\textbf{79.64\%}$ \\%
            
            \rowcolor{gray!15}
            FlowNet2 &
            $\textbf{79.92\%}$ &
            $\textbf{32/65}$ & 
            $79.51\%$ \\%
            
            \hline%
        \end{tabular}
    }

    \end{center}
    \caption{Motion segmentation and action recognition using different methods; see text for details. \textbf{Motion Segmentation:} We report results using \cite{Ochs14,KB15b} on the training set of \mbox{FBMS-59}~\cite{Bro10c,Ochs14} with a density of $4$ pixels. Different densities and error measures are given the supplemental material. ``\textit{Extracted objects}'' refers to objects with $\text{F}\geq 75\%$. $^\ddagger$FlowNetS is evaluated on 28 out of 29 sequences; on the sequence \emph{lion02}, the optimization did not converge even after one week. 
    \textbf{Action Recognition:} We report classification accuracies after training the temporal stream of     ~\cite{twostream}. We use a stack of 5 optical flow fields as input. Due to long training times only selected methods could be evaluated. $^\dagger$To reproduce results from \cite{twostream}, for action recognition we use the OpenCV LDOF implementation. 
     Note the generally large difference for FlowNetS and FlowNet2 and the performance compared to traditional methods.
    \label{tab:results_moseg_actreg}
    }
\end{table} 



\section{Conclusions} 
\label{sec:conclusion}

We have presented several improvements to the FlowNet idea that have led to accuracy that is fully on par with state-of-the-art methods while FlowNet 2.0 runs orders of magnitude faster. We have quantified the effect of each contribution and showed that all play an important role. The experiments on motion segmentation and action recognition show that the estimated optical flow with FlowNet 2.0 is reliable on a large variety of scenes and applications. The FlowNet 2.0 family provides networks running at speeds from 8 to 140fps. This further extends the possible range of applications. While the results on Middlebury indicate imperfect performance on subpixel motion, FlowNet 2.0 results highlight very crisp motion boundaries, retrieval of fine structures, and robustness to compression artifacts. Thus, we expect it to become the working horse for all applications that require accurate and fast optical flow computation. 


\section*{Acknowledgements} 

We acknowledge funding by the ERC Starting Grant VideoLearn, the DFG Grant BR-3815/7-1, and the EU Horizon2020 project TrimBot2020. 

{\small
\bibliographystyle{ieee}
\bibliography{egbib}
}

\cleardoublepage

\twocolumn[
\vspace*{30pt}
\Large \bf{
\begin{tabular}{c}
Supplementary Material for \\
"FlowNet 2.0: Evolution of Optical Flow Estimation with Deep Networks"\\
\end{tabular} 
}
\vspace*{65pt}
]

\setcounter{page}{1}

\setcounter{section}{0} 
\setcounter{figure}{0}
\setcounter{table}{0} 
\setcounter{equation}{0} 


\maketitle

\section{Video} 

Please see the supplementary video for FlowNet2 results on a number of diverse video sequences, a comparison between FlowNet2 and state-of-the-art methods, and an illustration of the speed/accuracy trade-off of the FlowNet 2.0 family of models.

\begin{figure}
\begin{center}
\includegraphics[width=0.4\linewidth]{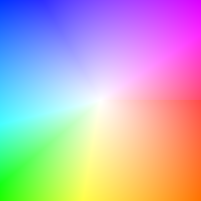}
\end{center}
\caption{Flow field color coding used in this paper. The displacement of every pixel in this illustration is the vector from the center of the square to this pixel. The central pixel does not move. The value is scaled differently for different images to best visualize the most interesting range. }
\label{fig:flow_key}
\end{figure}

\paragraph{Optical flow color coding.} For optical flow visualization we use the color coding of Butler \etal~\cite{sintel}.
The color coding scheme is illustrated in Figure~\ref{fig:flow_key}. 
Hue represents the direction of the displacement vector, while the intensity of the color represents its magnitude.
White color corresponds to no motion.
Because the range of motions is very different in different image sequences, we scale the flow fields before visualization: 
independently for each image pair shown in figures, and independently for each video fragment in the supplementary video.
Scaling is always the same for all methods being compared.

\section{Dataset Schedules: KITTI2015 Results} 

In Table~\ref{tab:dataset_schedules} we show more results of training networks with the original FlowNet schedule $\Sshort$~\cite{DFIB15} and the new FlowNet2 schedules $\Slong$ and $\Sfine$. We provide the endpoint error when testing on the KITTI2015 train dataset. Table 1 in the main paper shows the performance of the same networks on Sintel. One can observe that on KITTI2015, as well as on Sintel, training with $\Slong+\Sfine$ on the combination of \chairs and \things works best (in the paper referred to as \chairsToThings schedule). 

\begin{table}[t]
  \begin{center}
  {\small
  \setlength{\tabcolsep}{0.1cm}
  \begin{tabular}{|c|c||c||cc|} 
  \hline
  Architecture                       & Datasets               & $\Sshort$ & $\Slong$ & $\Sfine$ \\
  
  \hline \hline 
  
  \multirow{5}{*}{FlowNetS}   & \chairs                & $15.58$      & -          &  -         \\
                              & \chairs                 & -           & $14.60$     &  $14.28$    \\
                              & \things                & -           & $16.01$     &  $16.10$    \\
                              & mixed                  & -           & $16.69$     &  $15.57$    \\
                              & \chairsToThings    & -           & $14.60$     &  $\mathbf{14.18}$    \\
                              
  \hline \hline 
  
  \multirow{2}{*}{FlowNetC}   & \chairs                & $13.41$      & -          &  -         \\
                              & \chairsToThings    & -           & $12.48$     & \cellcolor{gray!15}$\mathbf{11.36}$    \\ 
  
  \hline
  \end{tabular}
  }
  \end{center}
  \caption{
  Results of training FlowNets with different schedules on different datasets (one network per row). Numbers indicate endpoint errors on the KITTI2015 training dataset. 
  \label{tab:dataset_schedules}
  }
\end{table}

\section{Recurrently Stacking Networks with the Same Weights} 

The bootstrap network differs from the succeeding networks by its task (it needs to predict a flow field from scratch) and inputs (it does not get a previous flow estimate and a warped image). The network after the bootstrap network only refines the previous flow estimate, so it can be applied to its own output recursively. We took the best network from Table~2 of the main paper and applied \netb recursively multiple times. We then continued training the whole stack with multiple \netb. The difference from our final FlowNet2 architecture is that here the weights are shared between the stacked networks, similar to a standard recurrent network. Results are given in Table \ref{tab:shared-weights}. In all cases we observe no or negligible improvements compared to the baseline network with a single \netb .

\begin{table}
  \begin{center}
  \begin{tabular}{|l||c|c|c|}%
  \hline
        & Training of    & Warping  &  \\
        & \netb          & gradient & EPE  \\
        & enabled        & enabled  &  \\
  \hline \hline
  \neta + $1\times$\netb & \mn  & --   & $\textbf{2.93}$      \\ 
  \neta + $2\times$\netb & \mn  & --   & $2.95$ \\
  \neta + $3\times$\netb & \mn  & --   & $3.04$ \\
  \hline 
  \neta + $3\times$\netb & \my & \mn  & $\textbf{2.85}$ \\ 
  \neta + $3\times$\netb & \my & \my  & $2.85$ \\
  \hline
  \end{tabular}
  \end{center}
  \caption{
  Stacked architectures using shared weights. The combination in the first row corresponds to the best results of Table~2 from the paper. Just applying the second network multiple times does not yield improvements. In the two bottom rows we show the results of fine-tuning the stack of the top networks on \chairs for 100k more iterations. This leads to a minor improvement of performance. 
  } \label{tab:shared-weights}
\end{table}

\section{Small Displacements}

\subsection{The ChairsSDHom Dataset} 


\begin{figure}
\centering
\resizebox{\linewidth}{!}{%
    \setlength{\tabcolsep}{1.6pt}%
        \begin{tabular}{ccc}%
            \includegraphics[width=0.3\linewidth]{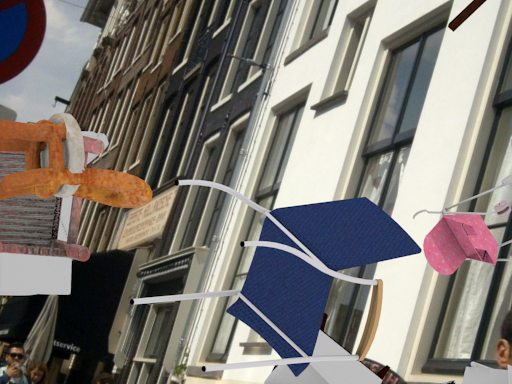} &
            \includegraphics[width=0.3\linewidth]{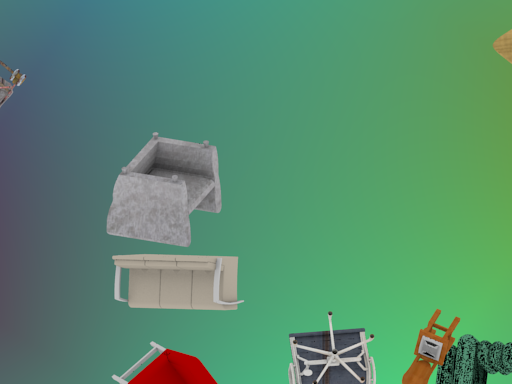} &
            \includegraphics[width=0.3\linewidth]{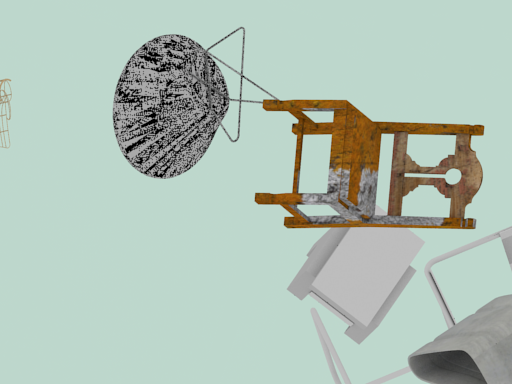} \\
        \end{tabular}%
}%
\caption{Images from the ChairsSDHom (Chairs Small Displacement Homogeneous) dataset.}
\label{fig:small_displacement_chairs}
\end{figure}

As an example of real-world data we examine the UCF101 dataset~\cite{ucf101}. We compute optical flow using LDOF~\cite{DFIB15} and compare the flow magnitude distribution to the synthetic datasets we use for training and benchmarking, this is shown in Figure~\ref{fig:displacements_histograms_plot}. 
While \chairs are similar to Sintel,
UCF101 is fundamentally different and contains much more small displacments.

To create a training dataset similar to UCF101, following~\cite{DFIB15}, we generated our ChairsSDHom (Chairs Small Displacement Homogeneous)  dataset by randomly placing and moving chairs in front of randomized background images. However, we also followed Mayer \etal~\cite{MIFDB16} in that our chairs are not flat 2D bitmaps as in \cite{DFIB15}, but rendered 3D objects.
Similar to Mayer \etal, we rendered our data first in a ``raw'' version to get blend-free flow boundaries and then a second time with antialiasing to obtain the color images. 
To match the characteristic contents of the UCF101 dataset, we mostly applied small motions.
We added scenes with weakly textured background to the dataset, being monochrome or containing a very subtle color gradient.
Such monotonous backgrounds are not unusual in natural videos, but almost never appear in \chairs or \things.
A featureless background can potentially move in any direction (an extreme case of the aperture problem), so we kept these background images fixed to introduce a meaningful prior into the dataset.
Example images from the dataset are shown in Figure~\ref{fig:small_displacement_chairs}. 


\begin{figure*}%
  \begin{center}%
    \input{displacements_histograms_plot}
  \end{center}%
  \caption{
      \textbf{Left:} histogram of displacement magnitudes of different datasets. y-axis is logarithmic. \textbf{Right:} zoomed view for very small displacements. The \chairs dataset very closely follows the Sintel dataset, while our \chairsSD datasets is close to UCF101. \things has few small displacements and for larger displacements also follows Sintel and \chairs. The \things histogram appears smoother because it contains more raw pixel data and due to its randomization of 6-DOF camera motion.
  }%
  \label{fig:displacements_histograms_plot}%
\end{figure*}%

\subsection{Fine-Tuning FlowNet2-CSS-ft-sd}

With the new \chairsSD dataset we fine-tuned our \FN{CSS} network for smaller displacements (we denote this by \textit{\mbox{\FN{CSS-ft-sd}}}). We experimented with different configurations to avoid sacrificing  performance on large displacements. We found the best performance can be achieved by training with mini-batches of $8$ samples: $2$ from \things and $6$ from \chairsSD. Furthermore, we applied a nonlinearity of $x^{0.4}$ to the endpoint error to emphasize the small-magnitude flows.

\begin{table}
    \begin{center}
  \resizebox{0.98\linewidth}{!}{%
  \begin{tabular}{|l|ccc|cc|c|}
      \hline
      Name     & Kernel         &Str. & Ch I/O      & In Res & Out  Res & Input\\
      \hline 
      conv0    & $3\!\times\!3$ & 1   & $\pz6/64$      & $512\!\times\!384$ & $512\!\times\!384$ & Images \\
      conv1    & $3\!\times\!3$ & 2   & $64/64$     & $512\!\times\!384$ & $256\!\times\!192$ & conv0   \\
      conv1\_1 & $3\!\times\!3$ & 1   & $\pz64/128$    & $256\!\times\!192$ & $256\!\times\!192$ & conv1   \\
      conv2    & $3\!\times\!3$ & 2   & $128/128$   & $256\!\times\!192$ & $128\!\times\!96\pz$  & conv1\_1 \\
      conv2\_1 & $3\!\times\!3$ & 1   & $128/128$   & $128\!\times\!96\pz$  & $128\!\times\!96\pz$  & conv2   \\
      conv3    & $3\!\times\!3$ & 2   & $128/256$   & $128\!\times\!96\pz$  & $64\!\times\!48$   & conv2\_1 \\
      conv3\_1 & $3\!\times\!3$ & 1   & $256/256$   & $64\!\times\!48$   & $64\!\times\!48$   & conv3   \\
      conv4    & $3\!\times\!3$ & 2   & $256/512$   & $64\!\times\!48$   & $32\!\times\!24$   & conv3\_1 \\
      conv4\_1 & $3\!\times\!3$ & 1   & $512/512$   & $32\!\times\!24$   & $32\!\times\!24$   & conv4   \\
      conv5    & $3\!\times\!3$ & 2   & $512/512$   & $32\!\times\!24$   & $16\!\times\!12$   & conv4\_1 \\
      conv5\_1 & $3\!\times\!3$ & 1   & $512/512$   & $16\!\times\!12$   & $16\!\times\!12$   & conv5   \\
      conv6    & $3\!\times\!3$ & 2   & $\pz512/1024$  & $16\!\times\!12$   & $8\!\times\!6$     & conv5\_1 \\
      conv6\_1 & $3\!\times\!3$ & 1   & $1024/1024$ & $8\!\times\!6$     & $8\!\times\!6$     & conv6   \\
      \hline
      pr6+loss6 & $3\!\times\!3$ & 1 & $1024/2\pz\pz\pz$    & $8\!\times\!6$     & $8\!\times\!6$     & conv6\_1 \\
      \hline
      upconv5   & $4\!\times\!4$ & 2 & $1024/512\pz$  & $8\!\times\!6$     & $16\!\times\!12$   & conv6\_1 \\
      rconv5    & $3\!\times\!3$ & 1 & $1026/512\pz$  & $16\!\times\!12$   & $16\!\times\!12$   & upconv5+pr6+conv5\_1 \\
      pr5+loss5 & $3\!\times\!3$ & 1 & $512/2\pz\pz$     & $16\!\times\!12$   & $16\!\times\!12$   & rconv5 \\
      upconv4   & $4\!\times\!4$ & 2 & $512/256$   & $16\!\times\!12$   & $32\!\times\!24$   & rconv5 \\
      rconv4    & $3\!\times\!3$ & 1 & $770/256$   & $32\!\times\!24$   & $32\!\times\!24$   & upconv4+pr5+conv4\_1 \\
      pr4+loss4 & $3\!\times\!3$ & 1 & $256/2\pz\pz$     & $32\!\times\!24$   & $32\!\times\!24$   & rconv4 \\
      upconv3   & $4\!\times\!4$ & 2 & $256/128$   & $32\!\times\!24$   & $64\!\times\!48$   & rconv4 \\
      rconv3    & $3\!\times\!3$ & 1 & $386/128$   & $64\!\times\!48$   & $64\!\times\!48$   & upconv3+pr4+conv3\_1 \\
      pr3+loss3 & $3\!\times\!3$ & 1 & $128/2\pz\pz$     & $64\!\times\!48$   & $64\!\times\!48$   & rconv3 \\
      upconv2   & $4\!\times\!4$ & 2 & $128/64\pz$    & $64\!\times\!48$   & $128\!\times\!96\pz$  & rconv3 \\
      rconv2    & $3\!\times\!3$ & 1 & $194/64\pz$    & $128\!\times\!96\pz$  & $128\!\times\!96\pz$  & upconv2+pr3+conv2\_1 \\
      pr2+loss2 & $3\!\times\!3$ & 1 & $64/2\pz$      & $128\!\times\!96\pz$  & $128\!\times\!96\pz$  & rconv2 \\
      \hline
    \end{tabular}}
  \end{center}
  \caption{The details of the FlowNet2-SD architecture.}
  \label{table:sd_net}
\end{table}

\begin{table}
    \begin{center}
  \resizebox{0.98\linewidth}{!}{%
  \begin{tabular}{|l|ccc|cc|c|}
      \hline
      Name     & Kernel         &Str. & Ch I/O      & In Res & Out  Res & Input\\
      \hline 
      conv0    & $3\!\times\!3$ & 1   & $\pz6/64$      & $512\!\times\!384$ & $512\!\times\!384$  & Img1+flows+mags+errs \\
      conv1    & $3\!\times\!3$ & 2   & $64/64$     & $512\!\times\!384$ & $256\!\times\!192$  & conv0   \\
      conv1\_1 & $3\!\times\!3$ & 1   & $\pz64/128$    & $256\!\times\!192$ & $256\!\times\!192$  & conv1   \\
      conv2    & $3\!\times\!3$ & 2   & $128/128$   & $256\!\times\!192$ & $128\!\times\!96\pz$   & conv1\_1 \\
      conv2\_1 & $3\!\times\!3$ & 1   & $128/128$   & $128\!\times\!96\pz$  & $128\!\times\!96\pz$   & conv2   \\
      \hline
      pr2+loss2 & $3\!\times\!3$ & 1 & $128/2\pz\pz$     & $128\!\times\!96\pz$  & $128\!\times\!96\pz$   & conv2\_1 \\
      \hline
      upconv1   & $4\!\times\!4$ & 2 & $128/32\pz$    & $128\!\times\!96\pz$  & $256\!\times\!192$  & conv2\_1 \\
      rconv1    & $3\!\times\!3$ & 1 & $162/32\pz$    & $256\!\times\!192$ & $256\!\times\!192$  & upconv1+pr2+conv1\_1 \\
      pr1+loss1 & $3\!\times\!3$ & 1 & $32/2\pz$      & $256\!\times\!192$ & $256\!\times\!192$  & rconv1 \\
      upconv0   & $4\!\times\!4$ & 2 & $32/16$     & $256\!\times\!192$ & $512\!\times\!384$ & rconv1 \\
      rconv0    & $3\!\times\!3$ & 1 & $82/16$     & $512\!\times\!384$ & $512\!\times\!384$ & upconv0+pr1+conv0 \\
      pr0+loss0 & $3\!\times\!3$ & 1 & $16/2\pz$      & $512\!\times\!384$ & $512\!\times\!384$ & rconv0 \\
      \hline
    \end{tabular}}
  \end{center}
  \caption{The details of the FlowNet2 fusion network architecture.}
  \label{table:fusion_net}
\end{table}

\subsection{Network Architectures} 

The architectures of the small displacement network and the fusion network are shown in Tables~\ref{table:sd_net} and~\ref{table:fusion_net}. The input to the small displacement network is formed by concatenating both RGB images, resulting in $6$ input channels. The network is in general similar to FlowNetS. Differences are the smaller strides and smaller kernel sizes in the beginning and the convolutions between the upconvolutions. 

The fusion network is trained to merge the flow estimates of two previously trained networks, and this task dictates the input structure. 
We feed the following data into the network: the first image from the image pair, two estimated flow fields, their magnitudes, and finally the two squared Euclidean photoconsistency errors, that is, per-pixel squared Euclidean distance between the first image and the second image warped with the predicted flow field. 
This sums up to $11$ channels. Note that we do not input the second image directly. All inputs are at full image resolution, flow field estimates from previous networks are upsampled with nearest neighbor upsampling.



\section{Evaluation} 

\subsection{Intermediate Results in Stacked Networks}

The idea of the stacked network architecture is that the estimated flow field is gradually improved by every network in the stack. This improvement has been quantitatively shown in the paper. Here, we additionally show qualitative examples which clearly highlight this effect. The improvement is especially dramatic for small displacements, as illustrated in Figure~\ref{fig:intermediate_sd}. The initial prediction of FlowNet2-C is very noisy, but is then significantly refined by the two succeeding networks. The FlowNet2-SD network, specifically trained on small displacements, estimates small displacements well even without additional refinement. Best results are obtained by fusing both estimated flow fields. Figure~\ref{fig:intermediate_ld} illustrates this for a large displacement case. 

\newcommand{\interWidth}{0.3\linewidth}%
\newcommand{\interHeight}{0.23\linewidth}%

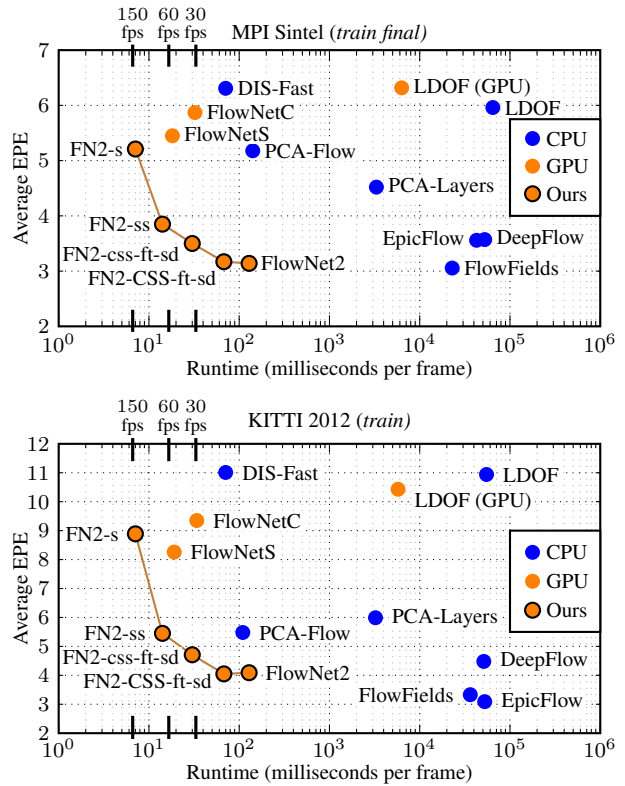
\begin{figure}[t]
  \begin{center}%
        
    \begin{tikzpicture}[%
      font=\footnotesize,%
      xscale=1.2,yscale=1.75*.42,%
      cpunode/.style={fill=blue,circle,scale=.6},%
      gpunode/.style={fill=orange,circle,scale=.6},%
      ournode/.style={fill=orange,draw,thick,circle,scale=.6},%
      l/.style={fill=white,inner sep=1pt}%
    ]%
      \draw[dotted] (0,0) grid (6,5);%
      \foreach \logbase in {0,1,...,5} {%
        \foreach \log in {.3,.48,.6,.7,.78,.85,.9,.95} {%
          \draw[dotted,black!30] ({\logbase+\log},0) -- ++(0,5);%
        }%
      }%
      \draw[thick] (0,0) -- ++(6,0) -- ++(0,5) -- ++(-6,0) -- cycle;%
      \node at (3,5.3) {MPI Sintel (\textit{train final)}};%
      \node[rotate=90] at (-.4,2.5) {Average EPE};%
      \node at (3,-.8) {Runtime (milliseconds per frame)};%
      \foreach \y in {2,3,...,7} {%
        \node at (0,{\y-2}) [anchor=east] {$\y$};%
        \draw (0,{\y-2}) -- ++(.05,0);%
        \draw (6,{\y-2}) -- ++(-.05,0);%
      }%
      \foreach \x in {0,1,...,6} {%
        \node at (\x,-.3) {$10^{\x}$};%
        \draw (\x,0) -- ++(0,.2);%
        \draw (\x,5) -- ++(0,-.2);%
      }%
      \foreach \logbase in {0,1,...,5} {%
        \foreach \log in {.3,.48,.6,.7,.78,.85,.9,.95} {%
          \draw ({\logbase+\log},0) -- ++(0,.1);%
          \draw ({\logbase+\log},5) -- ++(0,-.1);%
        }%
      }%
      \draw[thick,fill=white] (5,2) rectangle ++(1,1.75);%
      \node[cpunode] at (5.25,3.4) {};%
      \node at (5.3,3.4) [anchor=west] {CPU};%
      \node[gpunode] at (5.25,2.9) {};%
      \node at (5.3,2.9) [anchor=west] {GPU};%
      \node[ournode] at (5.25,2.4) {};%
      \node at (5.3,2.4) [anchor=west] {Ours};%
      
      \draw[very thick] (0.82,5.1) -- ++(0,-0.4);
      \draw[very thick] (0.82,-.1) -- ++(0,0.4);
      \node at (0.82,5.65) {\scriptsize $150$};
      \node at (0.82,5.3) {\scriptsize fps};
      \draw[very thick] (1.22,5.1) -- ++(0,-0.4);
      \draw[very thick] (1.22,-.1) -- ++(0,0.4);
      \node at (1.22,5.65) {\scriptsize $60$};
      \node at (1.22,5.3) {\scriptsize fps};
      \draw[very thick] (1.52,5.1) -- ++(0,-0.4);
      \draw[very thick] (1.52,-.1) -- ++(0,0.4);
      \node at (1.52,5.65) {\scriptsize $30$};
      \node at (1.52,5.3) {\scriptsize fps};
      
      \draw[brown,thick] (0.85,3.21)--(1.15,1.85)--(1.48,1.50)--(1.83,1.17)--(2.11,1.14);
      
      \node[cpunode] at (4.63,{3.558-2}) {};%
      \node[cpunode] at (4.72,{3.571-2}) {};%
      \node[cpunode] at (4.36,{3.055-2}) {};%
      \node[cpunode] at (4.81,{5.961-2}) {};%
      \node[gpunode] at (3.80,{6.318-2}) {};%
      \node[cpunode] at (2.15,{5.178-2}) {};%
      \node[cpunode] at (3.52,{4.520-2}) {};%
      \node[cpunode] at (1.85,{6.309-2}) {};%
      \node[gpunode] at (1.26,{5.45 -2}) {};%
      \node[gpunode] at (1.51,{5.87 -2}) {};%
      \node[ournode] at (0.85,{5.21 -2}) {};%
      \node[ournode] at (1.15,{3.85 -2}) {};%
      \node[ournode] at (1.48,{3.50 -2}) {};%
      \node[ournode] at (1.83,{3.17 -2}) {};%
      \node[ournode] at (2.11,{3.14 -2}) {};%
      \node[l]at({4.63-.1},{3.558-2})[anchor=east]{EpicFlow};%
      \node[l]at({4.72+.1},{3.571-2})[anchor=west]{DeepFlow};%
      \node[l]at({4.36+.1},{3.055-2})[anchor=west]{FlowFields};%
      \node[l]at({4.81+.1},{5.961-2})[anchor=west]{LDOF};%
      \node[l]at({3.80+.1},{6.318-2})[anchor=west]{LDOF (GPU)};%
      \node[l]at({2.15+.1},{5.178-2})[anchor=west]{PCA-Flow};%
      \node[l]at({3.52+.1},{4.520-2})[anchor=west]{PCA-Layers};%
      \node[l]at({1.85+.1},{6.309-2})[anchor=west]{DIS-Fast};%
      \node[l]at({1.26+.1},{5.45 -2+.04})[anchor=west]{FlowNetS};%
      \node[l]at({1.51+.1},{5.87 -2+.02})[anchor=west]{FlowNetC};%
      \node[l]at({0.85-.1},{5.21-2})[anchor=east]{FN2-s};%
      \node[l]at({1.15-.1},{3.85-2})[anchor=east]{FN2-ss};%
      \node[l]at({1.48-.11},{3.50-2.0})[anchor=north east]{FN2-css-ft-sd};%
      \node[l]at({1.83-.05},{3.17-2.1})[anchor=north east]{FN2-CSS-ft-sd};%
      \node[l]at({2.11+.1},{3.14-2})[anchor=west]{FlowNet2};%
    \end{tikzpicture}%
    
    \begin{tikzpicture}[%
      font=\footnotesize,%
      xscale=1.2,yscale=1.75*.22,%
      cpunode/.style={fill=blue,circle,scale=.6},%
      gpunode/.style={fill=orange,circle,scale=.6},%
      ournode/.style={fill=orange,draw,thick,circle,scale=.6},%
      l/.style={fill=white,inner sep=1pt}%
    ]%
      \draw[dotted] (0,0) grid (6,10);%
      \foreach \logbase in {0,1,...,5} {%
        \foreach \log in {.3,.48,.6,.7,.78,.85,.9,.95} {%
          \draw[dotted,black!30] ({\logbase+\log},0) -- ++(0,10);%
        }%
      }%
      \draw[thick] (0,0) -- ++(6,0) -- ++(0,10) -- ++(-6,0) -- cycle;%
      \node at (3,10.75) {KITTI 2012 (\textit{train)}};%
      \node[rotate=90] at (-.4,5) {Average EPE};%
      \node at (3,-1.4) {Runtime (milliseconds per frame)};%
      \foreach \y in {2,3,...,12} {%
        \node at (0,{\y-2}) [anchor=east] {$\y$};%
        \draw (0,{\y-2}) -- ++(.05,0);%
        \draw (6,{\y-2}) -- ++(-.05,0);%
      }%
      \foreach \x in {0,1,...,6} {%
        \node at (\x,-.5) {$10^{\x}$};%
        \draw (\x,0) -- ++(0,.3);%
        \draw (\x,10) -- ++(0,-.3);%
      }%
      \foreach \logbase in {0,1,...,5} {%
        \foreach \log in {.3,.48,.6,.7,.78,.85,.9,.95} {%
          \draw ({\logbase+\log},0) -- ++(0,.15);%
          \draw ({\logbase+\log},10) -- ++(0,-.15);%
        }%
      }%
      \draw[thick,fill=white] (5,3.5) rectangle ++(1,3.5);%
      \node[cpunode] at (5.25,6.25) {};%
      \node at (5.3,6.25) [anchor=west] {CPU};%
      \node[gpunode] at (5.25,5.25) {};%
      \node at (5.3,5.25) [anchor=west] {GPU};%
      \node[ournode] at (5.25,4.25) {};%
      \node at (5.3,4.25) [anchor=west] {Ours};%
      
      \draw[very thick] (0.82,10.2) -- ++(0,-0.8);
      \draw[very thick] (0.82,-.2) -- ++(0,0.8);
      \node at (0.82,11.3) {\scriptsize $150$};
      \node at (0.82,10.6) {\scriptsize fps};
      \draw[very thick] (1.22,10.2) -- ++(0,-0.8);
      \draw[very thick] (1.22,-.2) -- ++(0,0.8);
      \node at (1.22,11.3) {\scriptsize $60$};
      \node at (1.22,10.6) {\scriptsize fps};
      \draw[very thick] (1.52,10.2) -- ++(0,-0.8);
      \draw[very thick] (1.52,-.2) -- ++(0,0.8);
      \node at (1.52,11.3) {\scriptsize $30$};
      \node at (1.52,10.6) {\scriptsize fps};
      
      \draw[brown,thick] (0.85,6.89)--(1.15,3.45)--(1.48,2.71)--(1.83,2.05)--(2.11,2.09);
      
      \node[cpunode] at (4.72,{ 3.09-2}) {};%
      \node[cpunode] at (4.71,{ 4.48-2}) {};%
      \node[cpunode] at (4.56,{ 3.33-2}) {};%
      \node[cpunode] at (4.74,{10.94-2}) {};%
      \node[gpunode] at (3.76,{10.43-2}) {};%
      \node[cpunode] at (2.04,{ 5.48-2}) {};%
      \node[cpunode] at (3.51,{ 5.99-2}) {};%
      \node[cpunode] at (1.85,{11.01-2}) {};%
      \node[gpunode] at (1.28,{8.26 -2}) {};%
      \node[gpunode] at (1.53,{9.35 -2}) {};%
      \node[ournode] at (0.85,{8.89 -2}) {};%
      \node[ournode] at (1.15,{5.45 -2}) {};%
      \node[ournode] at (1.48,{4.71 -2}) {};%
      \node[ournode] at (1.83,{4.05 -2}) {};%
      \node[ournode] at (2.11,{4.09 -2}) {};%
      \node[l]at({4.72+.15},{ 3.09-2})[anchor=west]{EpicFlow};%
      \node[l]at({4.71+.15},{ 4.48-2})[anchor=west]{DeepFlow};%
      \node[l]at({4.56-.15},{ 3.33-2})[anchor=east]{FlowFields};%
      \node[l]at({4.74+.15},{10.94-2})[anchor=west]{LDOF};%
      \node[l]at({3.76+.15},{10.43-2})[anchor=north west]{LDOF (GPU)};%
      \node[l]at({2.04+.15},{ 5.48-2})[anchor=west]{PCA-Flow};%
      \node[l]at({3.51+.15},{ 5.99-2})[anchor=west]{PCA-Layers};%
      \node[l]at({1.85+.15},{11.01-2})[anchor=west]{DIS-Fast};%
      \node[l]at({1.28+.15},{8.26-2})[anchor=west]{FlowNetS};%
      \node[l]at({1.53+.15},{9.35-2})[anchor=west]{FlowNetC};%
      \node[l]at({0.85-.15},{8.89-2})[anchor=east]{FN2-s};%
      \node[l]at({1.15-.15},{5.45-2})[anchor=east]{FN2-ss};%
      \node[l]at({1.48-.11},{4.71-1.7})[anchor=north east]{FN2-css-ft-sd};%
      \node[l]at({1.83-.1},{4.05-1.9})[anchor=north east]{FN2-CSS-ft-sd};%
      \node[l]at({2.11+.15},{4.09-2})[anchor=west]{FlowNet2};%
      
    \end{tikzpicture}%
  \end{center}%
  \caption{Runtime vs. endpoint error comparison to the fastest existing methods with available code. The FlowNet2 family outperforms other methods by a large margin.}%
  \label{fig:benchmark_plots}%
\end{figure}%

\begin{figure*}
    \begin{center}%
      \resizebox{\linewidth}{!}{%
        \setlength{\tabcolsep}{1.6pt}%
        \begin{tabular}{|c||ccc|c|}%
            \hline
            \begin{tabular}{c} 
                Image Overlay \\ 
                \includegraphics[width=\interWidth,height=\interHeight]{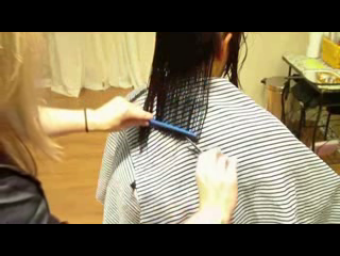}
            \end{tabular} & 
            \begin{tabular}{c} 
                FlowNet2-C output \\ 
                \includegraphics[width=\interWidth,height=\interHeight]{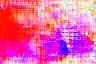}
            \end{tabular} & 
            \begin{tabular}{c} 
                FlowNet2-CS output \\ 
                \includegraphics[width=\interWidth,height=\interHeight]{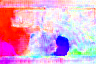}
            \end{tabular} & 
            \begin{tabular}{c} 
                FlowNet2-CSS output \\ 
                \includegraphics[width=\interWidth,height=\interHeight]{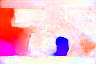}
            \end{tabular} &
            \multirow{2}{*}{
                \begin{tabular}{c} 
                    Fused output \\ 
                    \includegraphics[width=\interWidth,height=\interHeight]{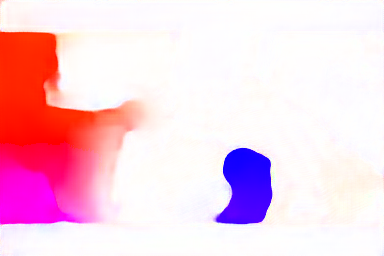}
                \end{tabular} 
            }
            \\
            \begin{tabular}{c} 
                FlowFields \\ 
                \includegraphics[width=\interWidth,height=\interHeight]{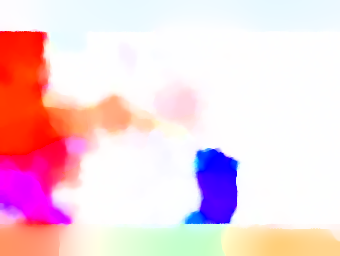}
            \end{tabular} & 
            &
            &
            \begin{tabular}{c} 
                FlowNet2-SD output \\ 
                \includegraphics[width=\interWidth,height=\interHeight]{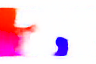}
            \end{tabular} & \\
            \hline
            \multicolumn{5}{c}{}
            \\ \hline 
            \begin{tabular}{c} 
                Image Overlay \\ 
                \includegraphics[width=\interWidth,height=\interHeight]{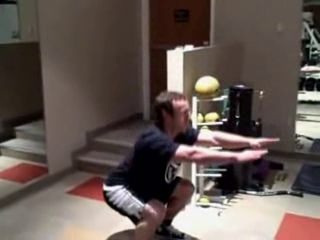}
            \end{tabular} & 
            \begin{tabular}{c} 
                FlowNet2-C output \\ 
                \includegraphics[width=\interWidth,height=\interHeight]{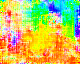}
            \end{tabular} & 
            \begin{tabular}{c} 
                FlowNet2-CS output \\ 
                \includegraphics[width=\interWidth,height=\interHeight]{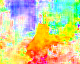}
            \end{tabular} & 
            \begin{tabular}{c} 
                FlowNet2-CSS output \\ 
                \includegraphics[width=\interWidth,height=\interHeight]{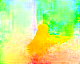}
            \end{tabular} &
            \multirow{2}{*}{
                \begin{tabular}{c} 
                    Fused output \\ 
                    \includegraphics[width=\interWidth,height=\interHeight]{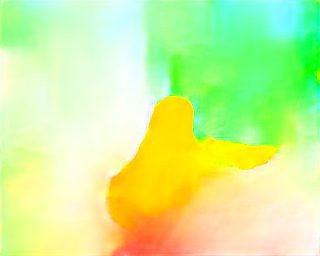}
                \end{tabular} 
            }
            \\
            \begin{tabular}{c} 
                FlowFields \\ 
                \includegraphics[width=\interWidth,height=\interHeight]{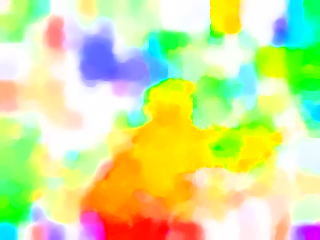}
            \end{tabular} & 
            &
            &
            \begin{tabular}{c} 
                FlowNet2-SD output \\ 
                \includegraphics[width=\interWidth,height=\interHeight]{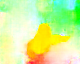}
            \end{tabular} & \\
            \hline 
            \multicolumn{5}{c}{}
            \\ 
            \hline 
            \begin{tabular}{c} 
                Image Overlay \\ 
                \includegraphics[width=\interWidth,height=\interHeight]{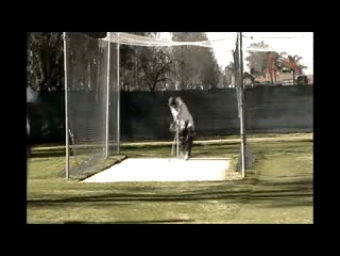}
            \end{tabular} & 
            \begin{tabular}{c} 
                FlowNet2-C output \\ 
                \includegraphics[width=\interWidth,height=\interHeight]{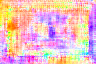}
            \end{tabular} & 
            \begin{tabular}{c} 
                FlowNet2-CS output \\ 
                \includegraphics[width=\interWidth,height=\interHeight]{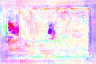}
            \end{tabular} & 
            \begin{tabular}{c} 
                FlowNet2-CSS output \\ 
                \includegraphics[width=\interWidth,height=\interHeight]{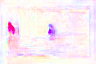}
            \end{tabular} &
            \multirow{2}{*}{
                \begin{tabular}{c} 
                    Fused output \\ 
                    \includegraphics[width=\interWidth,height=\interHeight]{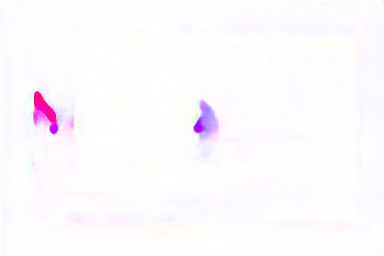}
                \end{tabular} 
            }
            \\
            \begin{tabular}{c} 
                FlowFields \\ 
                \includegraphics[width=\interWidth,height=\interHeight]{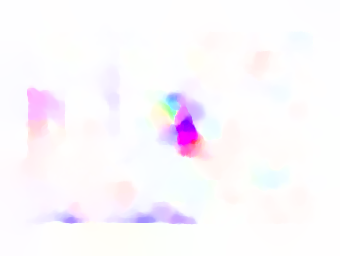}
            \end{tabular} & 
            &
            &
            \begin{tabular}{c} 
                FlowNet2-SD \\ 
                \includegraphics[width=\interWidth,height=\interHeight]{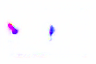}
            \end{tabular} & 
            \\ \hline
        \end{tabular}%
       }%
  \end{center}%
  \caption{Three examples of iterative flow field refinement and fusion for small displacements. The motion is very small (therefore mostly not visible in the image overlays). One can observe that FlowNet2-SD output is smoother than FlowNet2-CSS output. The fusion correctly uses the FlowNet2-SD output in the areas where FlowNet2-CSS produces noise due to small displacements.}%
  \label{fig:intermediate_sd}%
  \vspace*{-2mm}
\end{figure*}

\renewcommand{\interWidth}{0.3\linewidth}%
\renewcommand{\interHeight}{0.13\linewidth}%

\begin{figure*}
    \begin{center}%
      \resizebox{\linewidth}{!}{%
        \setlength{\tabcolsep}{1.6pt}%
        \begin{tabular}{|c||ccc|c|}%
            \hline
            \begin{tabular}{c} 
                Image Overlay \\ 
                \includegraphics[width=\interWidth,height=\interHeight]{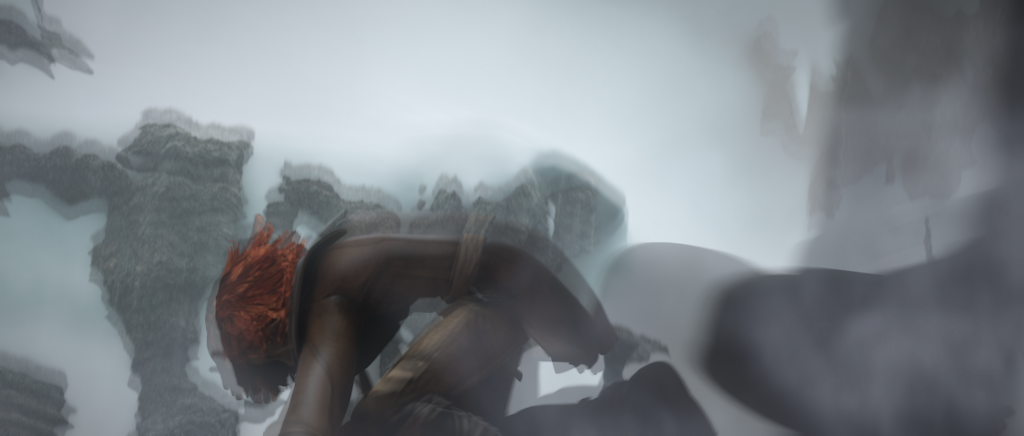}
            \end{tabular} & 
            \begin{tabular}{c} 
                FlowNet2-C output \\ 
                \includegraphics[width=\interWidth,height=\interHeight]{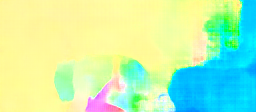}
            \end{tabular} & 
            \begin{tabular}{c} 
                FlowNet2-CS output \\ 
                \includegraphics[width=\interWidth,height=\interHeight]{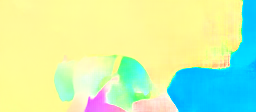}
            \end{tabular} & 
            \begin{tabular}{c} 
                FlowNet2-CSS output \\ 
                \includegraphics[width=\interWidth,height=\interHeight]{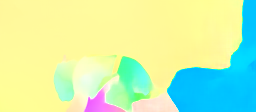}
            \end{tabular} &
            \multirow{2}{*}{
                \begin{tabular}{c} 
                    Fused output \\ 
                    \includegraphics[width=\interWidth,height=\interHeight]{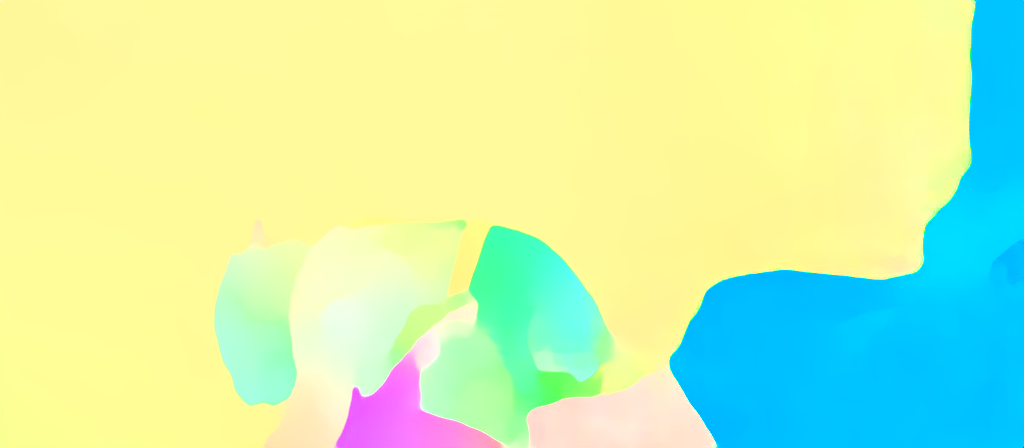}
                \end{tabular} 
            }
            \\
            \begin{tabular}{c} 
                Ground Truth \\ 
                \includegraphics[width=\interWidth,height=\interHeight]{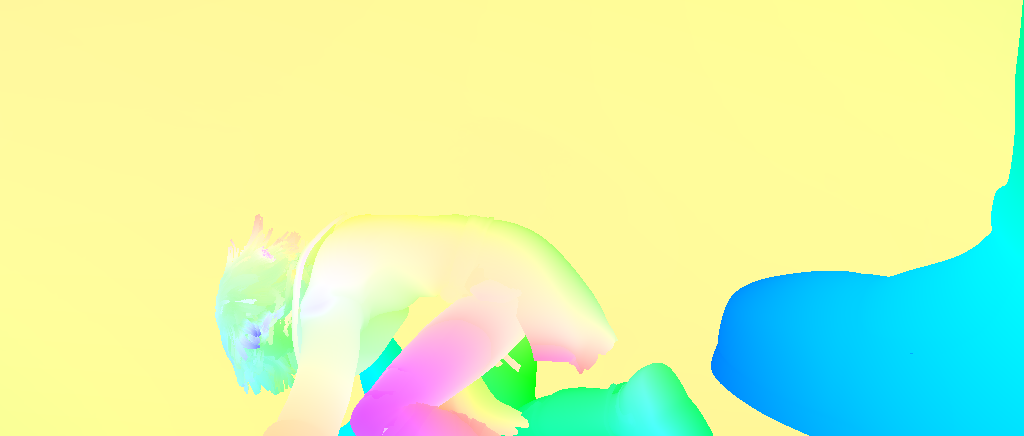}
            \end{tabular} & 
            &
            &
            \begin{tabular}{c} 
                FlowNet2-SD output \\ 
                \includegraphics[width=\interWidth,height=\interHeight]{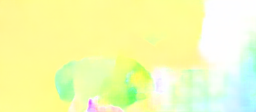}
            \end{tabular} & 
            \\
            \hline 
        \end{tabular}%
       }%
  \end{center}%
  \caption{Iterative flow field refinement and fusion for large displacements. The large displacements branch correctly estimates the large motions; the stacked networks improve the flow field and make it smoother. The small displacement branch cannot capture the large motions and the fusion network correctly chooses to use the output of the large displacement branch.}%
  \label{fig:intermediate_ld}%
\end{figure*}

\subsection{Speed and Performance on KITTI2012} 

Figure~\ref{fig:benchmark_plots} shows runtime vs. endpoint error comparisons of various optical flow estimation methods on two datasets: Sintel (also shown in the main paper) and KITTI2012. In both cases models of the FlowNet 2.0 family offer an excellent speed/accuracy trade-off. Networks fine-tuned on KITTI are not shown. The corresponding points would be below the lower border of the KITTI2012 plot.

\subsection{Motion Segmentation} 

Table~\ref{tab:results_motion_seg} shows detailed results on motion segmentation obtained using the algorithms from \cite{Ochs14,KB15b} with flow fields from different methods as input. For FlowNetS the algorithm does not fully converge after one week on the training set. Due to the bad flow estimations of FlowNetS~\cite{DFIB15}, only very short trajectories can be computed (on average about $3$ frames), yielding an excessive number of trajectories. Therefore we do not evaluate FlowNetS on the test set. On all metrics, FlowNet2 is at least on par with the best optical flow estimation methods and on the VI (variation of information) metric it is even significantly better.

\begin{table*}[tb]
  \begin{center}
  \resizebox{\textwidth}{!}{%
  \begin{tabular}{|l||cccccc|cccccc|}
    \hline%
    
    Method
    & \multicolumn{6}{c|}{\textbf{Training set} (29 sequences)}
    & \multicolumn{6}{c|}{\textbf{Test set} (30 sequences)} \\[0.5mm]
     & {D} & {P} & {R} & {F} & {VI} & {O}
     & {D} & {P} & {R} & {F} & {VI} & {O} \\
    \hline \hline 
    
    {LDOF (CPU) \cite{ldof}} &{0.81\%} & {86.73\%} & {73.08\%} & {79.32\%}& {0.267} & {31/65} & {0.87\%}& {87.88\%} &{67.70\%} & {76.48\%} & {0.366}  & {25/69} \\
    {DeepFlow \cite{deepflow}} &{\textbf{0.86\%}} & {88.96\%} & {\textbf{76.56\%}} & {\bf 82.29\%} & {0.296}& \textbf{33/65} & {0.89\%} &\textbf{88.20\%} & {69.39\%} & {\bf 77.67\%}  &  {0.367}&{26/69} \\
    {EpicFlow \cite{epicflow}} &{0.84\%} & {87.21\%} & {74.53\%}&  {80.37\%} &  {0.279}&{30/65} & \textbf{0.90\%} &{85.69\%} & {69.09\%} & {76.50\%}   & {0.373}&{25/69} \\
    {FlowFields \cite{flowfields}} &{0.83\%} & {87.19\%} & {74.33\%}  &{80.25\%}& {0.282} & {31/65} & {0.89\%} &{86.88\%} & \textbf{69.74\%} & {77.37\%}& {0.365}  & \textbf{27/69} \\
    {FlowNetS~\cite{DFIB15}} &{0.45\%}&{74.84\%} & {45.81\%} & {56.83\%} & {0.604} & {\pz3/65} &{0.48\%}&{68.05\%} & {41.73\%} & {51.74\%}  & {0.60}& {\pz3/69}\\
    {FlowNet2-css-ft-sd} &{0.78\%} & {88.07\%} & {71.81\%}  & {79.12\%}& {0.270} & {28/65} & {0.81\%} &{83.76\%} & {65.77\%} & {73.68\%} & {0.394} & {24/69}\\
    {FlowNet2-CSS-ft-sd} &{0.79\%} & {87.57\%} & {73.87\%}  & {80.14\%}& {0.255} & {31/65} & {0.85\%} &{85.36\%} & {68.81\%} & {76.19\%} & {0.327} & {26/69}\\
    \rowcolor{gray!15}
    {FlowNet2} &{0.80\%} & {\textbf{89.63\%}} & {73.38\%}  & {80.69\%} & {\bf 0.238}& {29/65} & {0.85\%} &{86.73\%} & {68.77\%} & {76.71\%} & {\bf 0.311} & {26/69}\\
    
    \hline
    
    {LDOF (CPU) \cite{ldof}} &{3.47\%} & {86.79\%} & {73.36\%}&  {79.51\%} &{0.270}& {28/65} & {3.72\%} &{86.81\%} & {67.96\%} & {76.24\%}  & {0.361}&{25/69} \\
    {DeepFlow \cite{deepflow}} &{\textbf{3.66\%}} & {86.69\%} & {\textbf{74.58\%}} & {\bf 80.18\%} & {0.303}& {29/65} & {3.79\%} &{88.58\%} & {68.46\%} & {\bf 77.23\%}& {0.393}  & \textbf{27/69} \\
    {EpicFlow \cite{epicflow}} &{3.58\%} & {84.47\%} & {73.08\%}&  {78.36\%} & {0.289}& {27/65} & \textbf{3.83\%} &{86.38\%} & \textbf{70.31\%} & {77.52\%} & {0.343} & \textbf{27/69} \\
    {FlowFields \cite{flowfields}} &{3.55\%} & {87.05\%} & {73.50\%}  &{79.70\%} & {0.293}& {30/65} & {3.82\%} &{88.04\%} & {68.44\%} & {77.01\%}& {0.397}  & {24/69} \\
    {FlowNetS~\cite{DFIB15}}$^*$ &{1.93\%}&{76.60\%} & {45.23\%} & {56.87\%}& {0.680}  & {\pz3/62} &{--}&{--} & {--} & {--}& {--}  & {\pz--/69}\\
    {FlowNet2-css-ft-sd} &{3.38\%} & {85.82\%} & {71.29\%}  & {77.88\%}& {0.297} & {26/65} & {3.53\%} &{84.24\%} & {65.49\%} & {73.69\%} & {0.369} & {25/69}\\
    {FlowNet2-CSS-ft-sd} &{3.41\%} & {86.54\%} & {73.54\%}  & {79.52\%}& {0.279} & {30/65} & {3.68\%} &{85.58\%} & {67.81\%} & {75.66\%} & {0.339} & \textbf{27/69}\\
    \rowcolor{gray!15}
    {FlowNet2} &{3.41\%} & {\textbf{87.42\%}} & {73.60\%}  & {79.92\%}& {\bf 0.249} & \textbf{32/65} & {3.66\%} &{\textbf{87.16\%}} & {68.51\%} & {76.72\%} & {\bf 0.324} & {26/69} \\
    \hline%
  \end{tabular}%
  }
  \end{center}%
  \caption{\label{tab:results_motion_seg} Results on the FBMS-59~\cite{Bro10c,Ochs14} dataset on training (\textbf{left}) and test set (\textbf{right}). Best results are highlighted in bold.
  \textbf{Top}: low trajectory density (8px distance), \textbf{bottom}: high trajectory density (4px distance).
  We report \textbf{D}: density (depending on the selected trajectory sparseness), \textbf{P}: average precision, \textbf{R}: average recall, \textbf{F}: F-measure, \textbf{VI}: variation of information (lower is better), and \textbf{O}: extracted objects with $\text{F}\geq 75\%$. ($^*$) FlownetS is evaluated on 28 out of 29 sequences. On the sequence \emph{lion02}, the optimization did not converge after one week. Due to the convergence problems we do not evaluate FlowNetS on the test set.}
\end{table*}

\subsection{Qualitative results on KITTI2015} 

\begin{figure*}[t]
  \begin{center}%
      \resizebox{\linewidth}{!}{%
    \setlength{\tabcolsep}{1.6pt}%
    \begin{tabular}{ccc}%
       Image Overlay &
       Ground Truth &
       FlowFields \cite{flowfields} \\ 
        \includegraphics[width=0.4\textwidth]{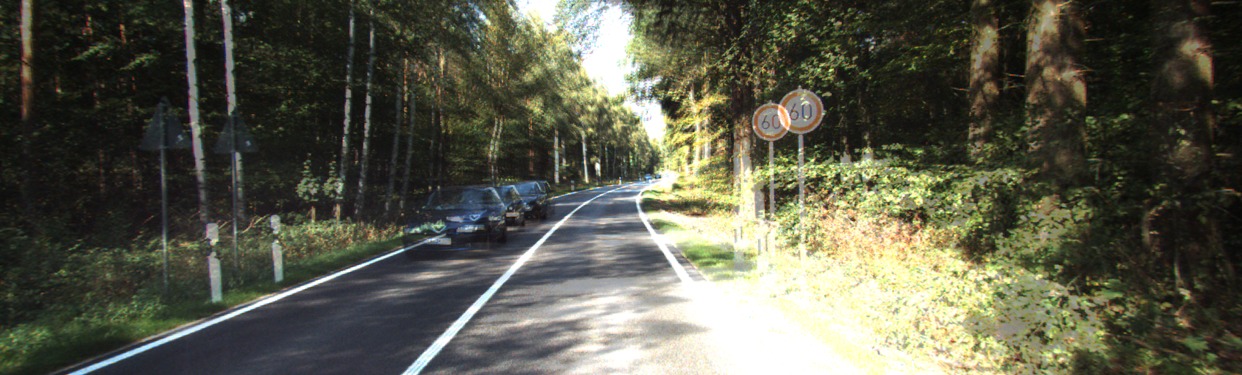} &
        \includegraphics[width=0.4\textwidth]{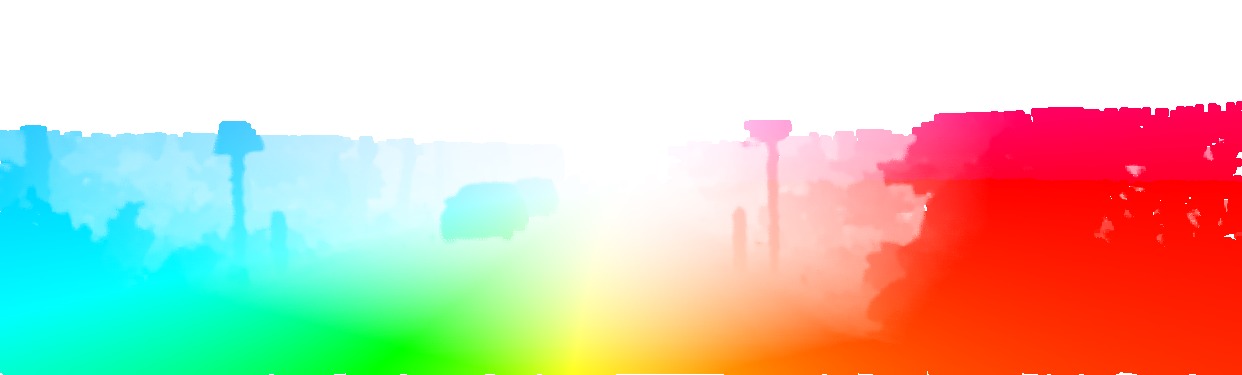} &
        \includegraphics[width=0.4\textwidth]{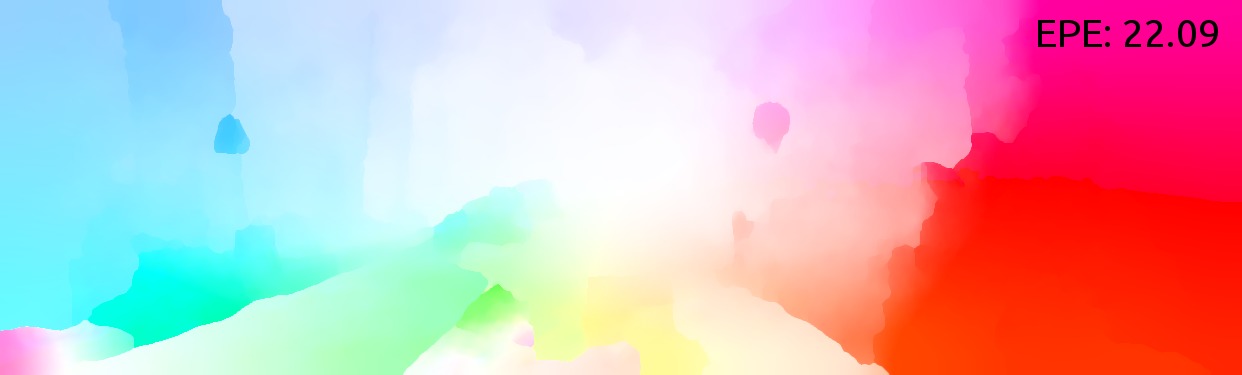} \\
       PCA-Flow \cite{pcaflowlayers}&
       FlowNetS \cite{DFIB15}&
       FlowNet2-kitti \\
        \includegraphics[width=0.4\textwidth]{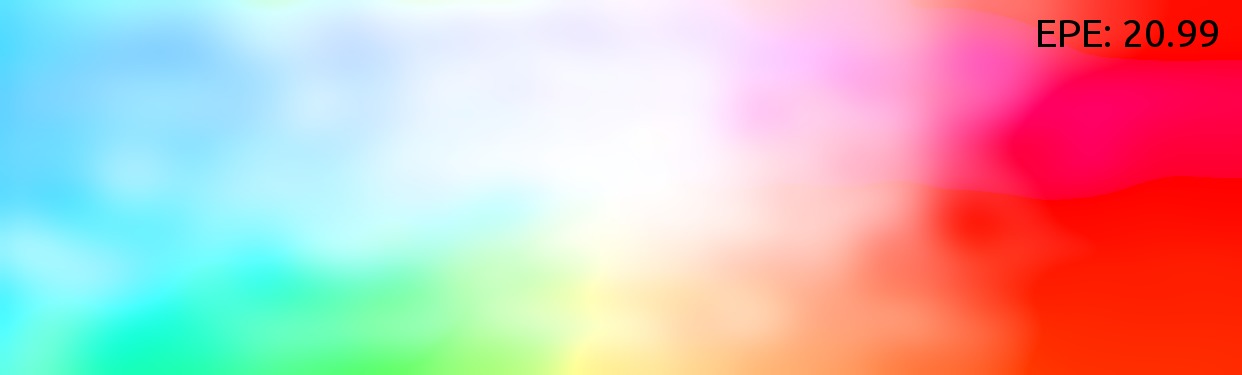} &
        \includegraphics[width=0.4\textwidth]{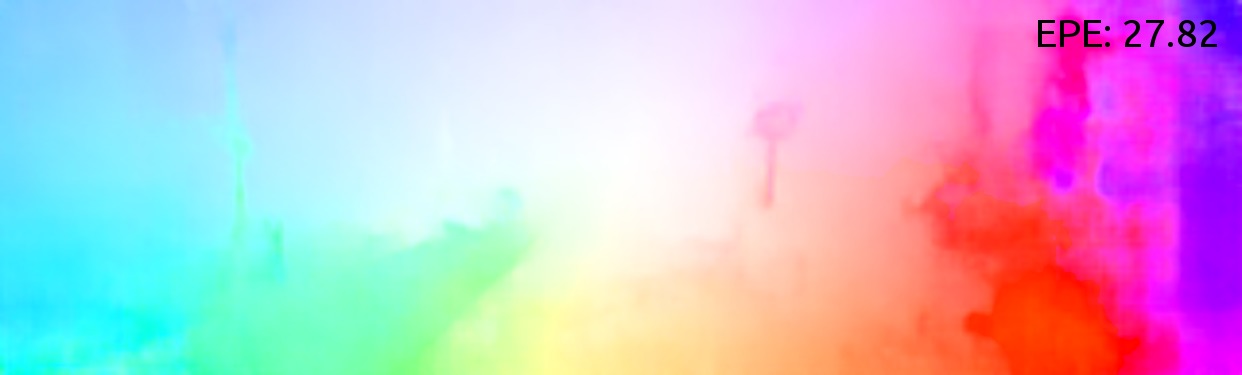} &        \includegraphics[width=0.4\textwidth]{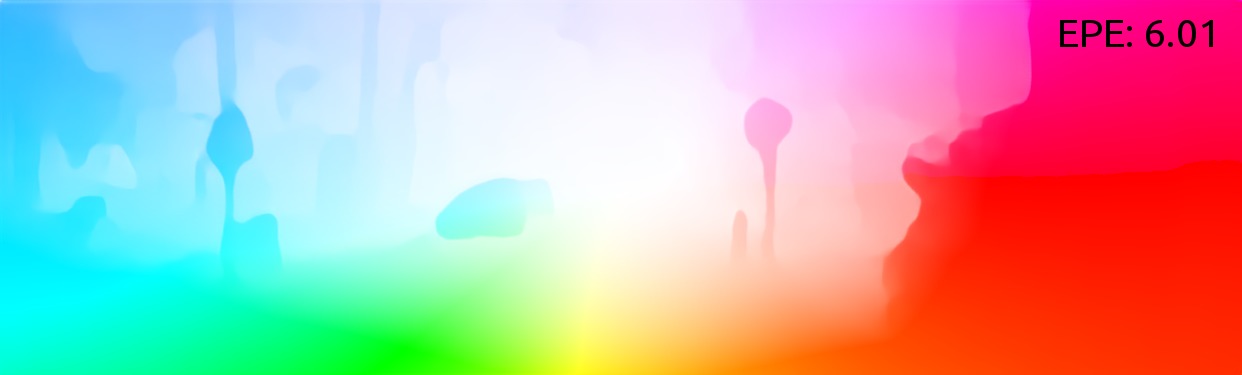} \\
    \hline%
    \end{tabular}%
   }%

\resizebox{\linewidth}{!}{%
    \setlength{\tabcolsep}{1.6pt}%
    \begin{tabular}{ccc}%
       Image Overlay &
       Ground Truth &
       FlowFields \cite{flowfields} \\ 
        \includegraphics[width=0.4\textwidth]{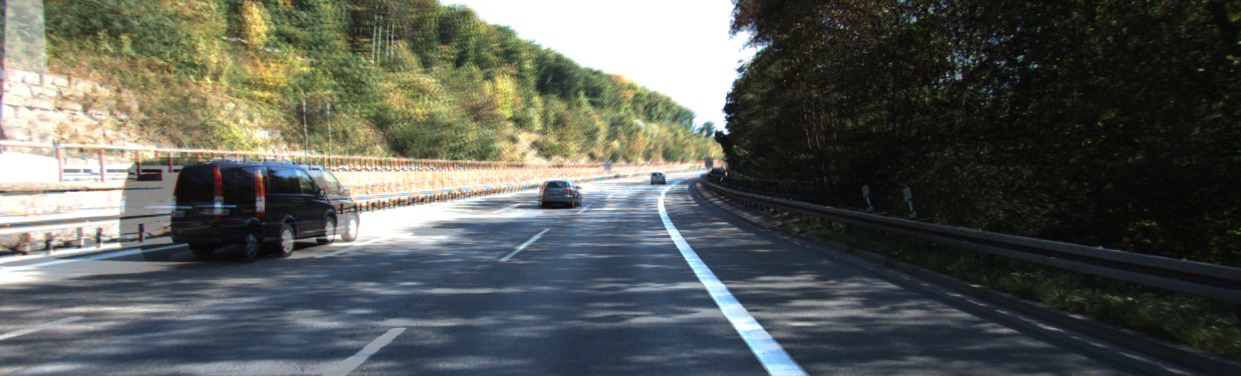} &
        \includegraphics[width=0.4\textwidth]{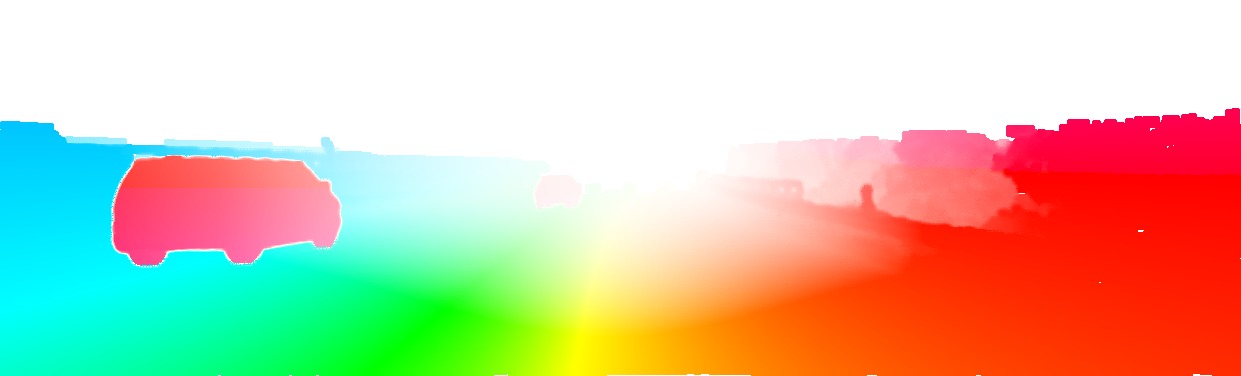} &
        \includegraphics[width=0.4\textwidth]{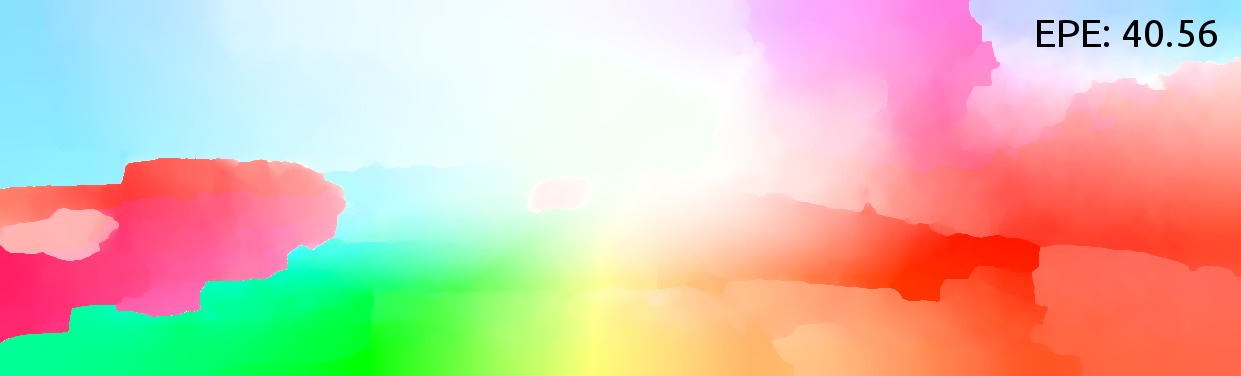} \\
       PCA-Flow \cite{pcaflowlayers}&
       FlowNetS \cite{DFIB15}&
       FlowNet2-kitti \\
        \includegraphics[width=0.4\textwidth]{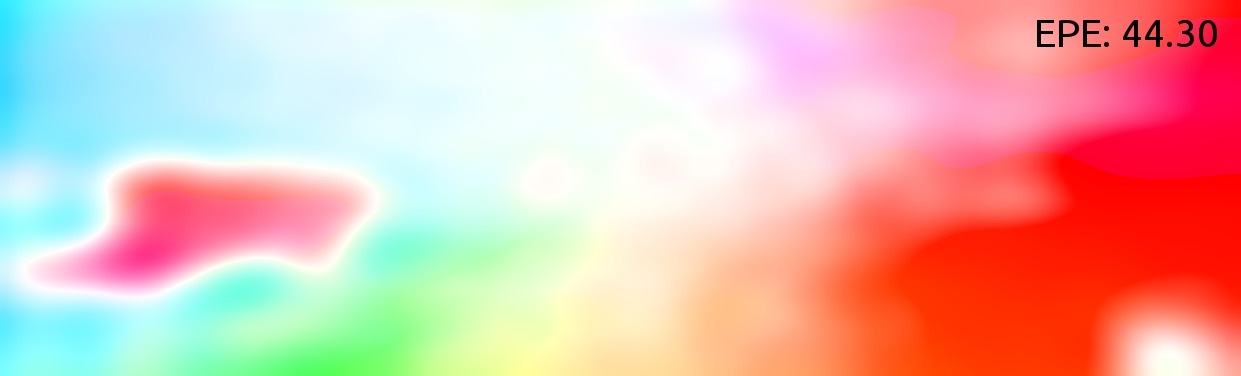} &
        \includegraphics[width=0.4\textwidth]{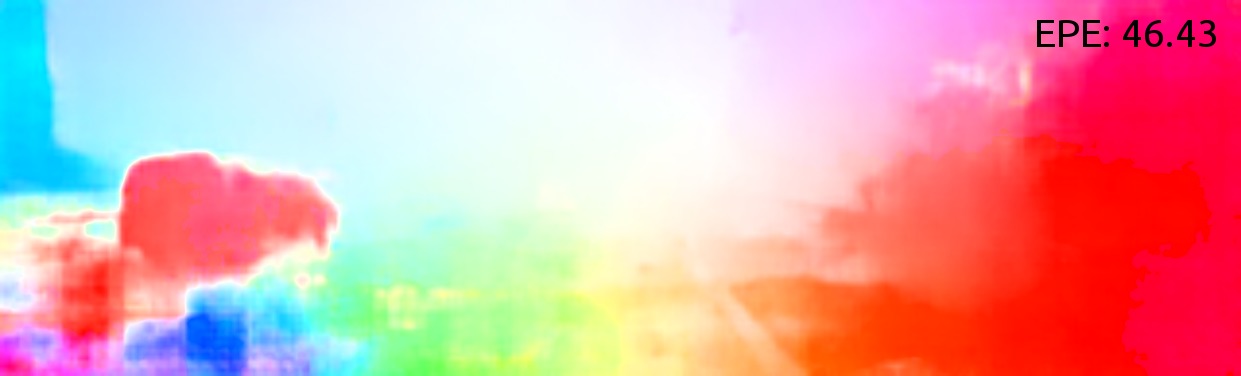} &        \includegraphics[width=0.4\textwidth]{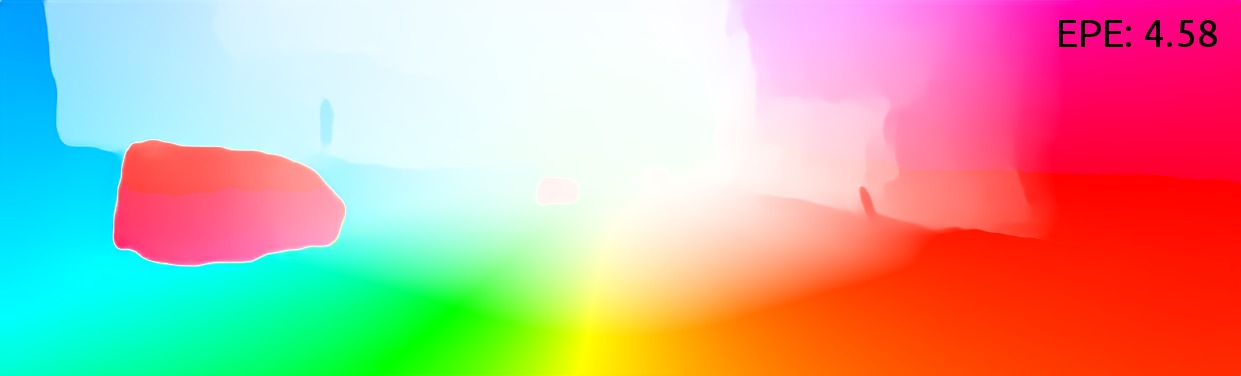} \\
    \end{tabular}%
   }%
 
  \end{center}%
  \caption{Qualitative results on the KITTI2015 dataset. Flow fields produced by FlowNet2-kitti are significantly more accurate, detailed and smooth than results of all other methods. Sparse ground truth has been interpolated for better visualization (note that this can cause blurry edges in the ground truth).}%
  \label{fig:gallery_kitti2015}%
\end{figure*}

Figure~\ref{fig:gallery_kitti2015} shows qualitative results on the KITTI2015 dataset. 
FlowNet2-kitti has not been trained on these images during fine-tuning.
KITTI ground truth is sparse, so for better visualization we interpolated the ground truth with bilinear interpolation. 
FlowNet2-kitti significantly outperforms competing approaches both quantitatively and qualitatively.



\renewcommand{\galleryWidth}{0.165\linewidth}%

\section{Warping Layer}

The following two sections give the mathematical details of forward and backward passes through the warping layer used to stack networks. 
\subsection{Definitions and Bilinear Interpolation} 

\newcommand*\conj[1]{\overline{#1}}

Let the image coordinates be $\mathbf{x}=(x,y)^\top$ and the set of valid image coordinates $R$. Let $\textbf{I}(\mathbf{x})$ denote the image and $\textbf{w}(\mathbf{x})=(u(\mathbf{x}),v(\mathbf{x}))^\top$ the flow field. The image can also be a feature map and have arbitrarily many channels. 
Let channel $c$ be denoted with $I_c(\mathbf{x})$. 
We define the coefficients:
\begin{eqnarray} 
  \theta_x = x-\Afloor{x} \mathbf{,\,\,\,\,\,\,\,\,} \conj{\theta_x} =  1-\theta_x  \mbox{,} \nonumber \\
  \theta_y = y-\Afloor{y} \mathbf{,\,\,\,\,\,\,\,\,} \conj{\theta_y} =  1-\theta_y  \mbox{\phantom{,}}
\end{eqnarray} 
and compute a continuous version $\tilde{\mathbf{I}}$ of $\mathbf{I}$ using bilinear interpolation in the usual way: 
\begin{equation}
  \begin{aligned}
    \tilde{\mathbf{I}}(x,y) 
    &=& \phantom{+\;} 
           \conj{\theta_x}\conj{\theta_y}\mathbf{I}(\Afloor{x}, \Afloor{y}) \\
    && +\; \theta_x\conj{\theta_y} \mathbf{I}(\Aceil{x}, \Afloor{y}) \\
    && +\; \conj{\theta_x}\theta_y \mathbf{I}(\Afloor{x}, \Aceil{y}) \\
    && +\; \theta_x\theta_y \mathbf{I}(\Aceil{x}, \Aceil{y})
  \end{aligned}
\end{equation}

\subsection{Forward Pass}

During the forward pass, we compute the warped image by following the flow vectors. We define all pixels to be zero where the flow points outside of the image:

\begin{equation}
\mathbf{J}_{\mathbf{I},\mathbf{w}}(\mathbf{x}) = 
\begin{cases}
\tilde{\mathbf{I}}(\mathbf{x} + \mathbf{w}(\mathbf{x})) & \mbox{ if } \mathbf{x} + \mathbf{w}(\mathbf{x}) \mbox{ is in } R \mbox{,} \\
0  & \mbox{ otherwise.} \\
\end{cases} 
\end{equation} 

\subsection{Backward Pass} 

During the backward pass, we need to compute the derivative of $\mathbf{J}_{\mathbf{I},\mathbf{w}}(\mathbf{x})$ with respect to its inputs $\mathbf{I}(\mathbf{x}')$ and  $\mathbf{w}(\mathbf{x}')$, where $\mathbf{x}$ and $\mathbf{x}'$ are different integer image locations.
Let $\delta(b) = 1$ if $b$ is true and 0 otherwise, and let $\mathbf{x}+\mathbf{w}(\mathbf{x}) = (p(\mathbf{x}), q(\mathbf{x}))^\top$. For brevity, we omit the dependence of $p$ and $q$ on $\mathbf{x}$. The derivative with respect to $I_c(\mathbf{x}')$ is then computed as follows: 
\begin{eqnarray} 
\frac{\partial J_c(\mathbf{x})}{\partial I_c(\mathbf{x}')} 
& = & \frac{\partial \tilde{I_c}(\mathbf{x}+\mathbf{w}(\mathbf{x}))}{\partial I_c(\mathbf{x}')} \nonumber  \\
& = & \frac{\partial \tilde{I_c}(p,q)}{\partial I_c(x',y')}  \nonumber \\
 & = &  \phantom{+\;} \conj{\theta_{x'}}\conj{\theta_{y'}}\delta(\Afloor{p} = x')\delta(\Afloor{q} = y') \nonumber \\
    && +\; \theta_{x'}\conj{\theta_{y'}} \delta(\Aceil{p} = x')\delta(\Afloor{q} = y') \nonumber  \\
    && +\; \conj{\theta_{x'}}\theta_{y'} \delta(\Afloor{p} = x')\delta(\Aceil{q} = y') \nonumber  \\
    && +\; \theta_{x'}\theta_{y'} \delta(\Aceil{p} = x')\delta(\Aceil{q} = y') \mbox{.}
\end{eqnarray} 

The derivative with respect to the first component of the flow $u(\mathbf{x})$ is computed as follows: 
\begin{eqnarray} 
\frac{\partial \mathbf{J}(\mathbf{x})}{\partial u(\mathbf{x'})} & = &  
\begin{cases}
0 & \mbox{if } \mathbf{x} \ne \mathbf{x'} \mbox{ or } (p, q)^\top \notin R \\
\frac{\partial \tilde{\mathbf{I}}(\mathbf{x}+\mathbf{w}(\mathbf{x}))}{\partial u(\mathbf{x})} & \mbox{otherwise.} 
\end{cases} 
\end{eqnarray} 

In the non-trivial case, the derivative is computed as follows: 
\begin{eqnarray} 
\frac{\partial \tilde{\mathbf{I}}(\mathbf{x}+\mathbf{w}(\mathbf{x}))}{\partial u(\mathbf{x})}
&=& \frac{\partial \tilde{\mathbf{I}}(p, q)}{\partial u} \nonumber \\
&=& \frac{\partial \tilde{\mathbf{I}}(p, q)}{\partial p} \nonumber \\
&=& \phantom{+\;} 
           \frac{\partial}{\partial p}\conj{\theta_p}\conj{\theta_q}\mathbf{I}(\Afloor{p}, \Afloor{q}) \nonumber \\
    && +\; \frac{\partial}{\partial p}\theta_p\conj{\theta_q} \mathbf{I}(\Aceil{p}, \Afloor{q}) \nonumber \\
    && +\; \frac{\partial}{\partial p}\conj{\theta_p}\theta_q \mathbf{I}(\Afloor{p}, \Aceil{q}) \nonumber \\
    && +\; \frac{\partial}{\partial p}\theta_p\theta_q \mathbf{I}(\Aceil{p}, \Aceil{q}) \nonumber \\
&=& -\;
           \conj{\theta_q}\mathbf{I}(\Afloor{p}, \Afloor{q}) \nonumber \\
    && +\; \conj{\theta_q} \mathbf{I}(\Aceil{p}, \Afloor{q}) \nonumber \\
    && -\; \theta_q \mathbf{I}(\Afloor{p}, \Aceil{q}) \nonumber \\
    && +\; \theta_q \mathbf{I}(\Aceil{p}, \Aceil{q}) \mbox{.}
\end{eqnarray} 
Note that the ceiling and floor functions ($\lceil\cdot\rceil$, $\lfloor\cdot\rfloor$) are non-differentiable at points with integer coordinates and we use directional derivatives in these cases. 
The derivative with respect to $v(\mathbf{x})$ is analogous.

\end{document}